\documentclass{article}
\usepackage{booktabs}

 \usepackage[nonatbib,preprint]{neurips_2026}

\usepackage{amsmath}
\usepackage{amssymb}
\usepackage{mathtools}
\usepackage{amsthm}
\usepackage{bbm}
\usepackage{media9}
\usepackage{siunitx}
\usepackage{pifont}
\usepackage{lipsum}
\usepackage{stfloats}
\usepackage{enumitem}
\usepackage[table,dvipsnames]{xcolor}
\usepackage{tcolorbox}
\usepackage{ragged2e}
\usepackage{wrapfig}
\usepackage{url}
\usepackage{hyperref}

\usepackage{caption}
\usepackage{subcaption}
\usepackage{outlines}
\usepackage{cuted}
\usepackage{colortbl}
\usepackage{bm}
\usepackage{array}
\usepackage{multirow}
\usepackage{tabularx}
\usepackage{algorithm}
\usepackage{algpseudocode}
\usepackage{tcolorbox}
\usepackage{makecell}

\newcommand\mypara[1]{\vspace{4pt}\noindent\textbf{#1.}}

\title{P2Voxel: Pyramid Pivot Voxelization for \\3D Mesh Tokenization}

\author{%
  Zhenhong Sun\thanks{Equal Contribution. Work done during an internship at Vertex Lab.} \\
  Australian National University \\
  Canberra, Australia \\
  \texttt{zhenhong.sun@anu.edu.au}
  \And
  Haozhe Liu\footnotemark[1] \\
  Vertex Lab \\
  Shanghai, China \\
  \texttt{l1583729854@gmail.com}
  \And
  Yifu Wang \\
  Vertex Lab \\
  Shanghai, China \\
  \texttt{1fwang927@gmail.com}
  \AND
  Xibin Song\thanks{Corresponding authors.} \\
  Vertex Lab \\
  Shanghai, China \\
  \texttt{song.sducg@gmail.com}
  \And
  Senbo Wang \\
  Vertex Lab \\
  Shanghai, China \\
  \texttt{wsb\_pro@live.com}
  \And
  Huadong Mo \\
  University of New South Wales \\
  Canberra, Australia \\
  \texttt{huadong.mo@unsw.edu.au}
  \AND
  Daoyi Dong\footnotemark[2] \\
  {\small University of Technology Sydney} \\
  Sydney, Australia \\
  \texttt{daoyidong@gmail.com}
  \And
  Hongdong Li \\
  {\small Australian National University} \\
  Canberra, Australia \\
  \texttt{hongdong.li@anu.edu.au}
  \And
  Pan Ji \\
  Vertex Lab \\
  Shanghai, China \\
  \texttt{peterji530@gmail.com}
}

\begin{document}
% Title portion
\maketitle

\begin{abstract}
Triangle meshes provide explicit and accurate surface geometry, yet their irregular topology connectivity makes 3D mesh tokenization a geometric sampling problem: how to sample and organize geometric evidence into compact, structured and learnable tokens.
Beyond field-centric volumetric sampling and edge-intersection surface sampling, we retarget mesh tokenization as \textit{local surface evidence sampling}: identifying the minimal geometric evidence inside each active voxel that is sufficient for deterministic surface recovery. 
To this end, we introduce \textbf{P2Voxel}, a pyramid pivot voxelization framework for compact and reconstruction-aware mesh tokenization. 
P2Voxel is built on three key innovations. 
Under the \textit{Local Planarity} assumption, Pivot Voxelization represents each active voxel with a surface pivot and an orientation sign, providing minimal local evidence that can induce the corner values required for deterministic reconstruction. 
Under the \textit{Spatial Complexity} assumption, Pyramid Pivot Voxelization exploits the spatial non-uniformity of real surfaces by allocating finer pivot tokens to geometrically complex regions while keeping smooth regions coarse and compact. 
Under the \textit{Block Reconstructability} assumption, a Pyramid VAE learns compact multi-resolution latent codes over locally reconstructable pivot blocks, avoiding the need to model the entire high-resolution voxelized shape as a dense global field. 
Together, these designs convert meshes into compact, structured, and learnable pyramid pivot tokens, enabling efficient mesh reconstruction for downstream 3D tasks.
Project page are released at \href{https://engineeringai-lab.github.io/pyramid-pivot-voxelization}{here}.
\end{abstract}

\section{Introduction}
Triangle meshes are the standard representation for 3D surfaces in graphics and geometry processing~\cite{foley1996computer,shirley2009fundamentals,yang2021geometry}, as they provide explicit and editable geometry for rendering, simulation, and downstream manipulation~\cite{akenine2019real,saito1990comprehensible,zangi2004water}. 
Yet their irregular connectivity makes them poorly suited as direct inputs or targets for modern 3D representation learning and generative models~\cite{zhang20233dshape2vecset,xiang2025structuredtrellis1,xiang2025nativetrellis2}, which favor compact, structured, and learnable tokens. 
This leads to the problem of \emph{3D mesh tokenization}: representing a complex surface by a finite set of geometric units that can be compressed, predicted, or generated. 
Rather than viewing tokenization as merely discretizing a shape, we view it as a problem of selecting the geometric evidence that should be preserved. 
Voxelization offers a natural spatial organization for such evidence by assigning local surface information to grid cells. 
The central challenge is therefore to design cell-wise tokens that are compact and regular, while still containing sufficient local surface evidence for deterministic high-quality mesh reconstruction.

Existing mesh tokenization methods can be viewed through the type of geometric evidence they preserve, including occupancy values~\cite{wu20153d}, point samples~\cite{nichol2022pointe}, implicit fields~\cite{mildenhall2021nerf,gao2022nerf}, signed distance fields (SDF)~\cite{oleynikova2016signedsdf}, and contouring constraints~\cite{ju2002dualDC}. 
For deterministic mesh reconstruction, grid-based SDFs and Dual Contouring represent two influential paradigms. 
Grid SDF methods~\cite{oleynikova2016signedsdf,jones20063d} store signed distances on spatial grids and extract surfaces through isosurface reconstruction, termed as \textit{field-centric volumetric sampling}. 
Although this sampling produces regular tokens, it encodes a two-dimensional surface indirectly through scalar samples in 3D space, requiring redundant storage for near-surface points even in sparse variants~\cite{deng2025efficientoctrees,ju2002dualhashgrids}. 
Dual Contouring~\cite{ju2002dualDC} follows an \textit{edge-intersection surface sampling} paradigm, using edge crossings and Hermite normals as boundary-level constraints or reconstruction rules that guide subsequent dual-vertex placement.
However, from a tokenization perspective, these edge-intersection samples are tied to sign-changing grid edges and mainly provide reconstruction cues for dual-vertex estimation, rather than defining a fixed and compact cell-wise token for the local surface patch itself.
We therefore ask whether mesh tokenization can move beyond volumetric field samples and edge-tied Hermite evidence, and instead ~\textit{represent each active voxel by the minimal local surface evidence required for deterministic reconstruction}.

Beyond field-centric volumetric sampling and edge-intersection surface sampling, we try to formulate mesh tokenization as \emph{local surface evidence sampling}: identifying the minimal geometric evidence inside each active voxel that is sufficient for deterministic surface recovery. 
Our first observation is \textit{local planarity}: at sufficiently high resolution, the surface patch inside a small active voxel can be approximated by a local plane. 
This observation is closely aligned with Marching Cubes~\cite{lorensen1998marching,dyken2008paralelmc}, where the surface topology and edge intersections within a voxel are determined by the signed scalar values at its corner vertices. 
However, if the local surface patch is approximated by a plane, these corner signs and distances can be derived from the plane geometry rather than stored independently.
This motivates our \textbf{Pivot Voxelization}, where each surface-intersecting voxel is encoded by a pivot point on the surface and an orientation sign, which together define the local plane and induce the corner values required by Sparse Marching Cubes.
Our second observation is \textit{spatial complexity}: real surfaces are highly non-uniform, with smooth regions that can be represented coarsely and sharp features, thin structures, or high-curvature details that require finer spatial support. 
This motivates \textbf{Pyramid Pivot Voxelization}, which keeps each pivot token simple while adaptively allocating finer pivot voxels only to geometrically complex regions. 
Our third observation is \textit{block reconstructability}: mesh recovery does not require learning the entire high-resolution voxelized shape as one monolithic object, because each pyramid block contains sufficient local surface evidence for deterministic reconstruction within its spatial extent. 
This motivates our \textbf{Pyramid VAE}, which learns compact multi-resolution latent codes over pyramid pivot blocks instead of consuming a dense global field. 
Together, these components form \textbf{P2Voxel}, a flexible framework that converts meshes into compact, structured, and learnable pivot tokens for scalable 3D mesh reconstruction.

The main contributions of this paper are summarized as follows:
\begin{itemize}[leftmargin=*, noitemsep, nolistsep]
    \item[$\bullet$] We formulate mesh tokenization as \textbf{local surface evidence sampling} and propose Pivot Voxelization, where a surface pivot and an orientation sign compactly represent the mesh.

    \item[$\bullet$] We introduce \textbf{Pyramid Pivot Voxelization}, which exploits spatial complexity by adaptively refining geometrically complex regions while keeping each pivot token simple and reconstructible.

    \item[$\bullet$] We then propose a \textbf{Pyramid VAE} based on block reconstructability, learning compact multi-resolution latent codes over locally pyramid pivot blocks for scalable mesh reconstruction.
\end{itemize}

\begin{figure}[t]
    \centering
    \includegraphics[width=0.95\textwidth]{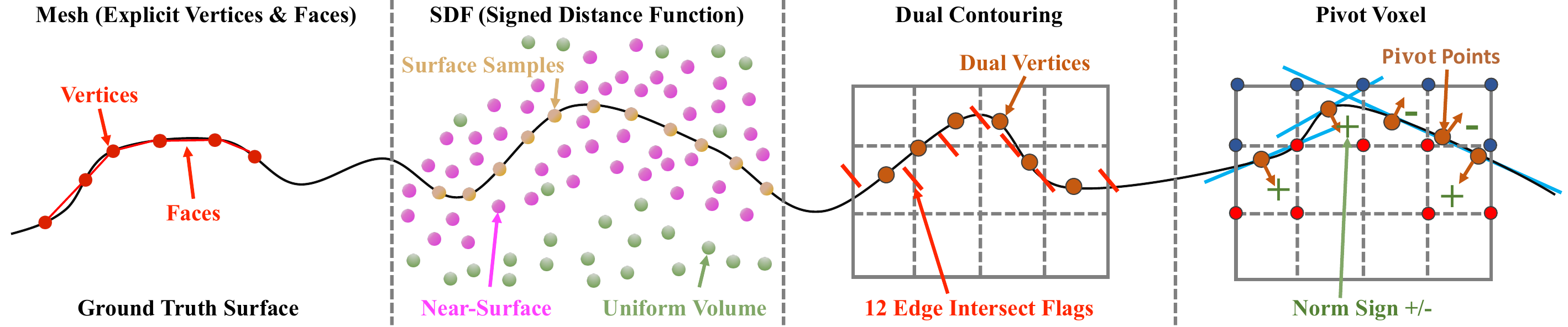}
    \caption{\textbf{Geometry Sampling for Mesh Tokenization}. 
    (a) Meshes are explicit but irregular. 
    (b) Grid SDFs use field-centric volumetric sampling by storing signed distance values. 
    (c) Dual Contouring uses edge-intersection surface sampling with edge crossings and Hermite constraints. 
    (d) Our Pivot Voxel uses local surface evidence sampling, with a pivot point and orientation sign.}
    \label{fig:teaser}
\end{figure}

\section{Related Work}
\mypara{Explicit 3D Representations for Generation}
Explicit representations, such as point clouds and meshes, are naturally discrete, making them intuitive candidates for Transformer tokenization. 
Point-based methods like Point-E~\cite{nichol2022pointe} and Shap-E~\cite{jun2023shap-e} generate sparse point clouds efficiently, but lack topological connectivity and often require post-processing such as Poisson reconstruction, which may fail on thin structures or sharp edges. 
Direct mesh generation methods, such as PolyGen~\cite{nash2020polygen} and MeshGPT~\cite{siddiqui2024meshgpt}, predict vertices and faces directly, but the combinatorial complexity of arbitrary mesh topology often limits them to specific categories or low-complexity shapes. 
Building on MeshAnything~\cite{chen2024meshanything}, MeshAnything V2~\cite{Chen_2025_ICCV} introduces Adjacent Mesh Tokenization (AMT) to reuse vertices across faces and shorten mesh sequences, while BPT~\cite{weng2025scaling} further improves scalability through block-wise indexing and patch aggregation for higher-resolution mesh generation.

\mypara{Implicit and Hybrid Representations}
Implicit representations, such as Neural Radiance Fields (NeRF)~\cite{mildenhall2021nerf} and Signed Distance Fields (SDF)~\cite{oleynikova2016signedsdf}, model geometry as continuous neural fields, while set-based approaches such as VecSet~\cite{zhang20233dshape2vecset} represent shapes as collections of surface elements. 
Recent generative models like TRELLIS~\cite{xiang2025structuredtrellis1} and DORA~\cite{chen2025dora} further employ structured latent codes, such as SLAT, to encode high-fidelity geometry. 
However, these methods usually rely on heavy neural decoders, typically MLPs, to extract explicit surfaces, introducing inference latency and often over-smoothing high-frequency details. 
Hybrid representations, such as FlexiCubes~\cite{zhao2024flexidreamerflexicubes} and Deep Marching Tetrahedra (DMTet)~\cite{shen2021deepDMTet}, bridge implicit and explicit representations, but still require extensive volumetric sampling beyond the actual surface to maintain field continuity.

\mypara{Sparse and Explicit Representations} Sparse data structures, such as Octrees \cite{deng2025efficientoctrees} and Hash Grids \cite{ju2002dualhashgrids}, are standard techniques for compressing voxel data. In geometry processing, Dual Contouring (DC)~\cite{ju2002dualDC} allows for sharp feature preservation by storing a single vertex per active voxel. However, maintaining valid topology in these frameworks often requires dense structural overhead. Notable recent generative approaches, such as FaithC~\cite{luo2025faithful} and TRELLIS 2~\cite{xiang2025nativetrellis2}, achieve high fidelity but rely on explicit edge-state encoding: they typically necessitate recording the intersection status across all 12 edges of a voxel to fully resolve topological ambiguities. This requirement increases the information density per token, diverting representational capacity toward structural bookkeeping.

\section{Methodology}
\label{sec:method}

\mypara{Problem Statement}
Triangle meshes are standard in graphics and geometry processing for their explicit and accurate surface description, but their irregular connectivity and non-uniform complexity make them poorly suited for compact tokenization in representation learning.
A central problem in geometric representation learning is to convert a surface mesh $\mathcal{M}$ into a \emph{compact} and \emph{structured} representation $\mathcal{T}$ that can be efficiently modeled while still allowing the accurate recovery of the original geometry.
An ideal representation should be (i) \emph{token-efficient} with as few elements as possible, (ii) \emph{structured} and easy to learn, and (iii) \emph{recoverable} by standard geometric reconstruction operators (e.g., Marching Cubes~\cite{lorensen1998marching}).
Formally, given a watertight mesh $\mathcal{M}$, we seek a token set $\mathcal{T}$ and a deterministic decoder $\mathcal{D}$ such that
\begin{equation}
\hat{\mathcal{M}}=\mathcal{D}(\mathcal{T}),
\qquad
|\mathcal{T}|\ \text{is minimized subject to}\ \hat{\mathcal{M}}\approx\mathcal{M}.
\end{equation}

To this end, we propose \textbf{P2Voxel}, a voxelization-based tokenization framework that converts a watertight mesh $\mathcal{M}$ into a compact and structured token set $\mathcal{T}$, while enabling accurate reconstruction with standard geometry operators.
P2Voxel contains three key components: Pivot Voxelization for sampling local surface evidence within active voxels, Pyramid Pivot Voxelization for adaptive multi-resolution token allocation, and a Pyramid VAE for learning compact latents over pivot blocks.

\begin{figure}[t]
    \centering
    \includegraphics[width=1\linewidth]{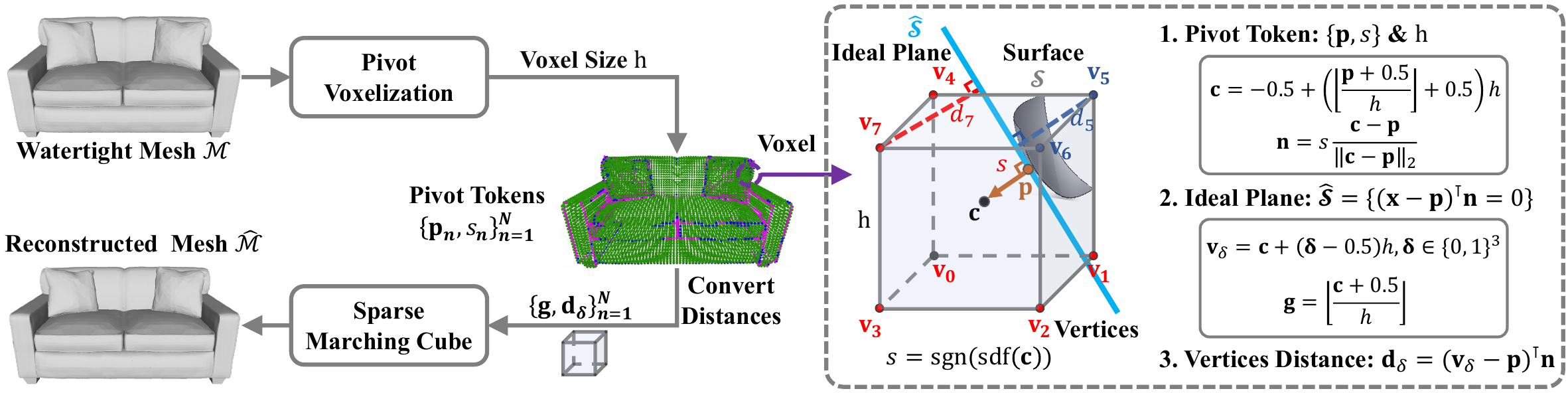}
    \caption{\textbf{Pivot Voxelization.} We voxelize a watertight mesh $\mathcal{M}$ with voxel size $h$ and represent each active voxel by a pivot token $\{\mathbf{p}, s\}$, where $\mathbf{p}$ is the pivot point and $s=\mathrm{sgn}(\mathrm{sdf}(\mathbf{c}))$ indicates the inside/outside orientation. An ideal plane $\hat{\mathcal{S}}$ can evaluate vertex distances $d_\delta$ for Sparse-MC.}
\label{fig:pivot_pipe1}
\end{figure}

\subsection{Pivot Voxelization}
\label{subsec:pivot_voxelization}

\begin{tcolorbox}[
  colback=gray!6,
  colframe=black!30,
  boxrule=0.5pt,
  arc=2pt,
  left=6pt,right=6pt,top=4pt,bottom=4pt
]
\noindent\textbf{Assumption 1 (Local Planarity).}
At sufficiently high resolution, each surface-intersecting voxel contains only a small local surface, which can be approximated by an \emph{ideal plane}.
\end{tcolorbox}

Sparse Marching Cubes (Sparse-MC)~\cite{lorensen1998marching,dyken2008paralelmc,tang2022cubvh} reconstructs watertight surfaces by extracting the zero-level set from signed distance values at the eight corners of surface-intersecting (\emph{active}) voxels. These corner values jointly encode the topology and geometry of the enclosed surface patch $\mathcal{S}$. As the voxelization resolution increases (i.e., $h\rightarrow 0$), $\mathcal{S}$ becomes asymptotically locally planar. We therefore adopt \textbf{Assumption~1 (Local Planarity)} and approximate the geometry within each active voxel by an \emph{ideal plane} $\mathcal{\hat{S}}$, which induces a consistent set of corner distances.

A plane in $\mathbb{R}^3$ is uniquely determined by a point and a normal direction. Given the voxelization prior, we encode each active voxel by a compact \emph{pivot token} $\{\mathbf{p}, s\}$, where $\mathbf{p}\in\mathbb{R}^3$ is a pivot point on the local surface patch and $s\in\{-1,+1\}$ is a binary inside/outside orientation cue. Concretely, $\mathbf{p}$ is obtained by projecting the voxel center $\mathbf{c}$ onto the mesh, i.e., the closest intersection point on $\mathcal{M}$, and $s$ is given by the center SDF sign, $s=\mathrm{sgn}(\mathrm{sdf}(\mathbf{c}))$.

Conversely, given $\{\mathbf{p}, s\}$ and voxel size $h$, we recover the voxel center $\mathbf{c}$ by snapping $\mathbf{p}$ to its containing voxel, then compute the oriented normal $\mathbf{n}$ and ideal plane $\hat{\mathcal{S}}$:
\begin{equation}
\mathbf{c}
= -\tfrac{1}{2}
+ \Big(\big\lfloor \tfrac{\mathbf{p}+0.5}{h}\big\rfloor + 0.5\Big)h .
\end{equation}
Then the oriented normal and the induced ideal plane are given by
\begin{equation}
\mathbf{n} = s\,\frac{\mathbf{c}-\mathbf{p}}{\|\mathbf{c}-\mathbf{p}\|_2},
\qquad
\hat{\mathcal{S}}
= \left\{\mathbf{x}\in\mathbb{R}^3 \ \middle|\ (\mathbf{x}-\mathbf{p})^\top \mathbf{n}=0 \right\}.
\end{equation}

With $(\mathbf{p},\mathbf{n})$, the eight voxel corners and their plane-induced signed distances are computed as:
\begin{equation}
\mathbf{v}_{\boldsymbol{\delta}}=\mathbf{c}+(\boldsymbol{\delta}-0.5)\,h,
\qquad
\boldsymbol{\delta}\in\{0,1\}^3,
\end{equation}
and the signed distance at each corner is given by the plane SDF
\begin{equation}
d_{\boldsymbol{\delta}} = \mathrm{SDF}(\mathbf{v}_{\boldsymbol{\delta}})
= (\mathbf{v}_{\boldsymbol{\delta}}-\mathbf{p})^\top \mathbf{n}.
\end{equation}
The eight samples $\{d_{\boldsymbol{\delta}}\}$ provide the per-voxel corner SDF values. Since a grid vertex is shared by neighboring voxels, independently predicted corner values may be inconsistent. We therefore enforce a \emph{vertex-consistent} scalar field by averaging predictions from incident active voxels $\Omega_\mathbf{v}$:
\begin{equation}
D(\mathbf{v}) = \frac{1}{|\Omega_\mathbf{v}|} \sum_{i \in \Omega_\mathbf{v}} (\mathbf{v} - \mathbf{p}_i)^\top \mathbf{n}_i,
\end{equation}
where $(\mathbf{p}_i, \mathbf{n}_i)$ denotes the pivot token parameters of the $i$-th voxel. Sparse-MC is then applied to the unified field $D(\cdot)$, ensuring shared corner values across adjacent voxels and producing a seamless watertight reconstruction $\hat{\mathcal{M}}$.
In summary, the pivot token provides a minimal encoding for stable distance evaluation and deterministic reconstruction. We discuss the feasibility conditions and practical constraints of pivot construction in \textbf{Appendix~\ref{app:pivot_feas}}.

\begin{figure}[t]
    \centering
    \includegraphics[width=1\linewidth]{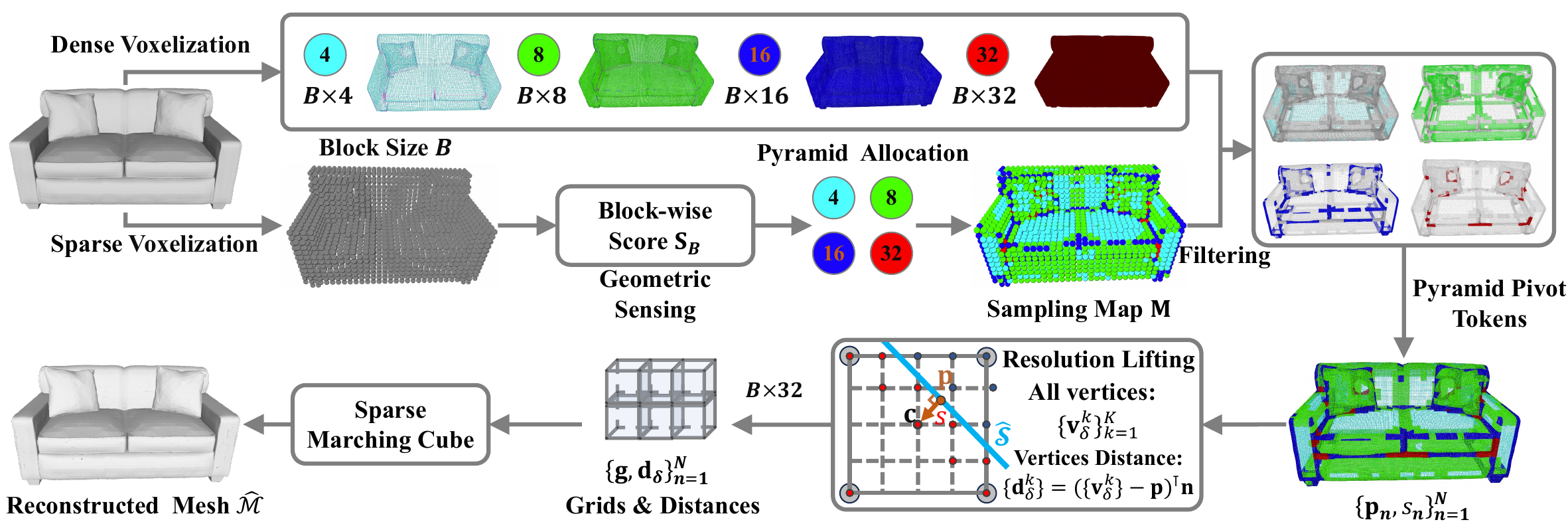}
    \caption{\textbf{Pyramid Pivot Voxelization.} We compute a block-wise score $s_B$ over macro-blocks of size $B$ to allocate adaptive sampling levels $\{4,8,16,32\}$, producing a pyramid sampling map $\mathbf{M}$. After filtering and resolution lifting, pyramid pivot tokens recover high-quality mesh.}
\label{fig:pivot_pipe2}
\end{figure}

\subsection{Pyramid Pivot Voxelization}
\label{subsec:pyramid_block}

\begin{tcolorbox}[colback=gray!5, colframe=gray!40, boxrule=0.5pt, sharp corners, left=4pt, right=4pt, top=4pt, bottom=4pt]
\noindent \textbf{Assumption 2 (Spatial Complexity).} 
\textit{Geometric information density is non-uniform: high-frequency details are concentrated around high-curvature, sharp, or thin structures, where fine-grained sampling is necessary, while coarse blocks suffice elsewhere.}
\end{tcolorbox}

Under \textbf{(Local Planarity)}, each active voxel admits a compact pivot representation, but this typically requires a sufficiently high resolution. 
Although a $512^3$ grid can capture the overall geometry with high fidelity, it remains inefficient due to redundant sampling in smooth regions. 
\textbf{Assumption~2 (Spatial Complexity)} motivates adaptive sampling with a block-wise pyramid strategy: fine-grained resolution is allocated only to high-frequency regions, while coarser blocks represent smooth areas. 

\mypara{Pyramid Sampling Map}
We partition the normalized space $[-0.5,0.5]^3$ into a macro-block grid
$\mathcal{B}\in\mathbb{Z}^{B\times B\times B}$ with block size $1/B$, and assign each block $\mathbf{g}$ a complexity score $\mathbf{S}_B(\mathbf{g})$.
Surface samples $\mathbf{x}$ are mapped to blocks by
$
\mathbf{g}(\mathbf{x})=\left\lfloor (\mathbf{x}+0.5)\,B \right\rfloor,
$
and curvature- and normal-based cues are aggregated as
\begin{equation}
\mathbf{S}_B(\mathbf{g})
=
\Big(
\bar{c}(\mathbf{g})
+\lambda_{\mathrm{var}}\,\sigma_c(\mathbf{g})
+\lambda_{\mathrm{grad}}\,\|\nabla \bar{c}(\mathbf{g})\|
+\lambda_{\mathrm{n}}\,\bar{\eta}(\mathbf{g})
\Big)\cdot \sqrt{N(\mathbf{g})},
\end{equation}
where $\bar{c}$ is the mean curvature magnitude, $\sigma_c$ is curvature variation, $\|\nabla \bar{c}\|$ captures spatial curvature change, $\bar{\eta}$ measures local normal deviation, and $N(\mathbf{g})$ is the number of surface samples in the block.
We discretize $\mathbf{S}_B$ into an allocation map
\begin{equation}
\mathbf{M}:\mathbf{g}\mapsto r \in \mathcal{R}=\{r_{min}, \dots, r_{max}\},
\end{equation}
where $\mathcal{R}$ is a predefined set of sampling resolutions with quantile-based binning over blocks, and $\mathbf{M}$ is restricted to surface-intersecting blocks, with missing active blocks filled by resolution $r_{min}$.

\mypara{Pyramid Sampling}
Given $\mathbf{M}:\mathbf{g}\mapsto r$, we perform pivot voxelization with $h_r = 1/r$, producing
\begin{equation}
\mathcal{T}_r=\Big\{\tau_n^{(r)}\Big\}_{n=1}^{N_r},
\qquad
\tau^{(r)}=\{\mathbf{p}^{(r)}, s^{(r)}\},
\end{equation}
where $\mathbf{p}^{(r)}\in\mathbb{R}^3$ is the pivot point and $s^{(r)}\in\{-1,+1\}$ is the orientation sign.
According to $\mathbf{M}$, for each active macro-block $\mathbf{g}$, tokens from resolution $r=\mathbf{M}(\mathbf{g})$ form the final pyramid set
\begin{equation}
\mathcal{T}_{\mathrm{pyr}}
=
\bigcup_{\mathbf{g}\in\mathcal{B}_{\mathrm{active}}}
\Big\{\tau^{(\mathbf{M}(\mathbf{g}))}\ \big|\ \tau \ \text{lies in block } \mathbf{g}\Big\}.
\end{equation}
Although constructing $\{\mathcal{T}_r\}_{r\in\mathcal{R}}$ is not a single-pass adaptive implementation, it is practical because closest-point queries and triangle normals are provided by optimized C++ backends~\cite{zhou2018open3d}.

\mypara{Pyramid Resolution Lifting}
Given $\mathcal{T}_{\mathrm{pyr}}$ over multiple resolutions $\mathcal{R}$, we lift all tokens to the maximum resolution $r_{\max}=\max(\mathcal{R})$ with voxel size $h_{\max}=1/r_{\max}$ for a unified Sparse-MC reconstruction.
For each token $\tau^{(r)}=\{\mathbf{p}^{(r)}, s^{(r)}\}\in\mathcal{T}_{\mathrm{pyr}}$, we recover its induced plane $\hat{\mathcal{S}}^{(r)}=(\mathbf{p}^{(r)},\mathbf{n}^{(r)})$, subdivide the original voxel into $\left({r_{\max}}/{r}\right)^3$ fine voxels, and evaluate their corner SDF samples to populate the sparse high-resolution field:
\begin{equation}
\mathbf{V}_{\max}=\{v_k\}_{k=1}^{K}, \qquad v_k \in \{0,\dots,r_{\max}-1\}^3,
\end{equation}
together with their corner-distance vectors
\begin{equation}
\mathbf{D}_{\max}=\Big\{\mathbf{d}(v_k)\Big\}_{k=1}^{K},
\qquad
\mathbf{d}(v_k)=\big[d_{\boldsymbol{\delta}}(v_k)\big]_{\boldsymbol{\delta}\in\{0,1\}^3}\in\mathbb{R}^{8},
\end{equation}
where each component is given by
\begin{equation}
d_{\boldsymbol{\delta}}(v_k)
= \big(\mathbf{v}_{k,\boldsymbol{\delta}}-\mathbf{p}^{(r)}\big)^\top \mathbf{n}^{(r)}.
\end{equation}
Finally, Sparse-MC is applied to $(\mathbf{V}_{\max},\mathbf{D}_{\max})$ to reconstruct the sparse watertight mesh $\hat{\mathcal{M}}$.

\begin{figure}[t]
    \centering
    \includegraphics[width=1\linewidth]{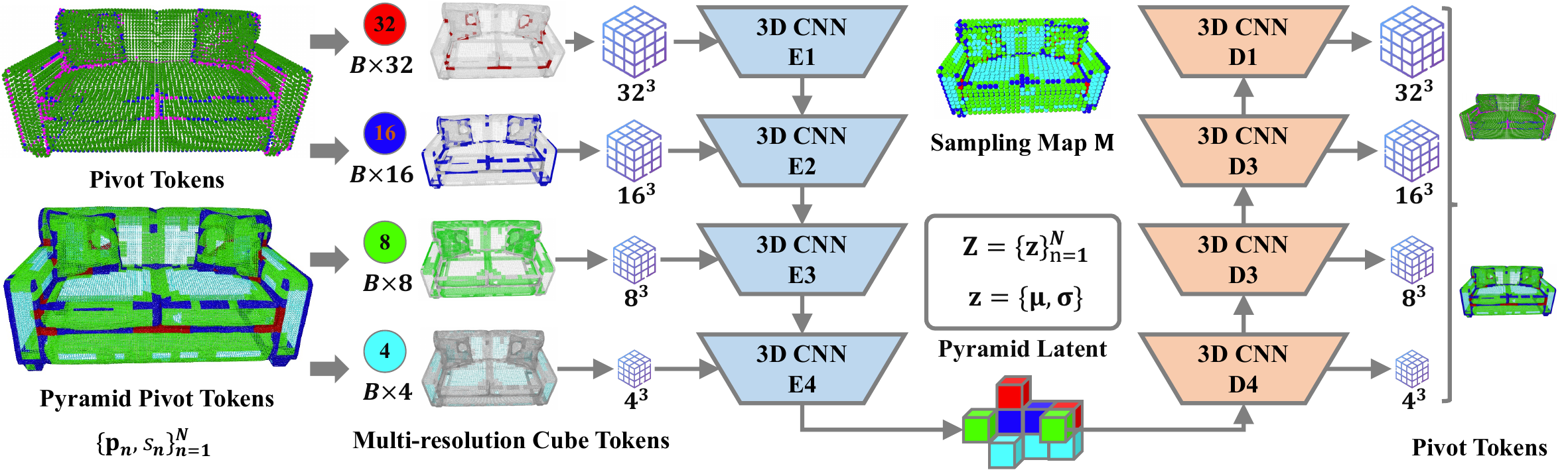}
    \caption{\textbf{Block-wise Pyramid VAE.} Given pivot tokens at adaptive sampling levels $\{4,8,16,32\}$, the encoder aggregates multi-resolution block features into a compact pyramid latent code, then reconstructs pivot tokens with the guidance of the sampling map $\mathbf{M}$.}
\label{fig:pivot_pipe2}
\end{figure}

\subsection{Pyramid VAE}
\label{subsec:pyramid_vae}

\begin{tcolorbox}[colback=gray!5, colframe=gray!40, boxrule=0.5pt, sharp corners, left=4pt, right=4pt, top=4pt, bottom=4pt]
\noindent \textbf{Assumption 3 (Block Reconstructability).} 
\textit{Surface reconstruction is locally composable: each pyramid block can recover its local surface patch from its pivot tokens, and the full mesh can be obtained by enforcing consistent corner values across neighboring blocks.}
\end{tcolorbox}

A straightforward strategy is to learn pivot tokens by decoding the entire shape on a dense global voxel grid. 
However, this quickly becomes prohibitive at high resolutions: even a minimal pivot field with three pivot coordinates and one orientation sign per voxel requires approximately $2.0$ GiB and $16.0$ GiB of FP32 memory at $512^3$ and $1024^3$ resolutions, respectively. 
In 3D CNN-based models, the actual training cost is much larger due to multi-channel feature maps, activations, gradients, and optimizer states. 
Thus, directly learning a global high-resolution voxel field is memory-inefficient and difficult to scale, even with sparse convolutional techniques~\cite{graham20183d}.

Based on \textbf{Assumption~3}, we instead design a \textbf{Pyramid VAE}. 
The key idea is that pivot tokens are locally reconstructable: each block contains sufficient surface evidence to recover its local patch through the induced plane and Sparse-MC corner values. 
Thus, the model does not need to encode the entire high-resolution shape as a monolithic dense tensor. 
Instead, we organize pyramid pivot tokens into multi-resolution block tensors according to the pyramid sampling map $\mathbf{M}$, where each level corresponds to a sampling resolution such as $\{4,8,16,32\}$ within macro-blocks of size $B$.

The encoder uses hierarchical 3D CNN branches to aggregate block-wise geometric features from coarse to fine levels and maps them into a shared pyramid latent representation
\[
\mathbf{z}=\{\mathbf{z}_n\}_{n=1}^{N},
\qquad
\mathbf{z}\sim q_\phi(\mathbf{z}\mid \mathcal{T}_{\mathrm{pyr}}, \mathbf{M}),
\]
where $\mathbf{M}$ provides the spatial allocation of sampling resolutions and $\mathcal{T}_{\mathrm{pyr}}$ provides the corresponding local surface evidence. 
The decoder mirrors this hierarchy to reconstruct pivot-token grids at multiple resolutions. 
After decoding, the predicted blocks are filtered, lifted to the maximum grid, and stitched through shared corner-value consistency for Sparse-MC reconstruction.

The pyramid representation separates \emph{where} to allocate resolution from \emph{what} surface evidence to store. 
It only adds a lightweight sampling map $\mathbf{M}$, which guides each active macro-block to reconstruct at its assigned resolution and can reduce to standard Pivot Voxelization when all blocks use the finest level. 
For generation, $\mathbf{M}$ can be learned as a coarse layout prior, while pivot tokens model local geometry, naturally supporting a coarse-to-fine generative process.

\section{Experiments}
\subsection{Implementation Details}
\mypara{Baselines}
We benchmark P2Voxel against representative baselines covering two mainstream paradigms: {(i) SDF-based sampling}, including {Vecset}~\cite{zhang20233dshape2vecset}, {Dora}~\cite{chen2025dora}, and {Hunyuan3D-2.1}~\cite{zhao2025hunyuan3d}; and {(ii) Dual Contouring}, represented by {FaithC}~\cite{luo2025faithful} and {TRELLIS~2}~\cite{xiang2025nativetrellis2}.
Our comparison focuses strictly on \emph{sampling strategy and surface reconstruction quality}, explicitly excluding downstream generative components (e.g., VAE tokenization or diffusion).
To ensure fairness, we standardize that all methods sample SDFs from identical watertight meshes and reconstruct using a neural implicit network~\cite{sitzmann2020implicit_siren}, which is detailed in \textbf{Appendix~\ref{app:detail}}.

\mypara{Datasets and Metrics}
We evaluate on three disjoint test sets ($\sim$400 shapes each) sampled from \textbf{ABO}~\cite{collins2022abo}, \textbf{Objaverse}~\cite{deitke2023objaverse}, and an in-the-wild (\textbf{Wild}) collection.
All shapes are preprocessed into watertight ground-truth surfaces using Dora's pipeline (UDF-to-SDF followed by Marching Cubes) to ensure topological consistency.
Quantitative metrics include Chamfer Distance (L1/L2), Earth Mover’s Distance (EMD)~\cite{rubner2000earth}, and F-score~\cite{sturm2012benchmark_fscore} ($\tau=0.002$).

\mypara{Setups}
We configure two voxelization strategies: a single-resolution baseline \textbf{Pivot-512} and our hierarchical Pyramid-R approach with $R_{max}=32$. 
We employ a 3-level hierarchy $\{128, 256, 512\}$ with sampling allocation ratios of $[0.85, 0.10, 0.05]$ for \textbf{Pyramid-512}, while extending to a 4-level hierarchy $\{128, 256, 512, 1024\}$ with ratios $[0.50, 0.35, 0.12, 0.03]$ for \textbf{Pyramid-1024}. 
Unless otherwise stated, hyperparameters are fixed at $\lambda_{var}=0.5$, $\lambda_{grad}=0.3$, and $n=3.0$. All experiments are conducted on a server with an AMD EPYC 7543 32-Core Processor and eight NVIDIA RTX 4090 GPUs (24GB VRAM). Detailed descriptions and runtime are provided in \textbf{Appendix~\ref{app:detail}}.

\begin{table*}[t]
\centering
\caption{\textbf{Quantitative evaluation of geometry reconstruction quality.}
We report Chamfer Distance (CDL1/CDL2), Earth Mover's Distance (EMD), and F-score ($\tau=0.002$) on ABO, Objaverse, and Wild datasets at \textbf{512} resolution (1024 Results in \textbf{Appendix}). We evaluate our three configurations: \textbf{Pivot-512 (single-scale $512$) against hierarchical Pyramid-512 ($128/256/512$) and Pyramid-1024 ($128/256/512/1024$)}.
Considering voxel dimension capacity and reconstruction fidelity, P-Voxel (Dim 4) achieves efficient performance, outperforming baselines and rivaling heavy-weight representations (e.g., FaithC with Dim 18). \textbf{Bold} indicates the best performance, and \underline{underlined} denotes the second best.
Qualitative results are provided in~\textbf{Figure~\ref{fig:quali_pyramid} of Appendix~\ref{app:qualitative}}.}
\label{tab:metrics_512}
\setlength{\tabcolsep}{5pt}
\resizebox{\textwidth}{!}{
\begin{tabular}{l l c c c c c c c c}
\toprule
\multirow{2}{*}{Dataset}
 & \multirow{2}{*}{Metrics}
 & \multicolumn{3}{c}{SDF} 
 & \multicolumn{2}{c}{Dual Contouring} 
 & \multicolumn{3}{c}{Ours} \\
\cmidrule(lr){3-5} \cmidrule(lr){6-7} \cmidrule(lr){8-10}
 & & Dora & Vecset & Hy3D & Trellis 2 & FaithC & Pivot-512 & Pyramid-1024 & Pyramid-512 \\
\midrule
Voxel & Dim & 1 & 1 & 1 & 7 & 18 & 4 & 4 & 4\\
\midrule
\multirow{5}{*}{ABO}
& Num $\downarrow$ & 731{,}072 & 500{,}000 & 624{,}640 & 908{,}296 & 820{,}445 & 880{,}134 & 385{,}151 & 135,736 \\
& CDL1 $\downarrow$ & 2.1858 & 2.2699 & 2.4972 & 2.1205 & \underline{2.1175} & \textbf{2.1166} & 2.1583 & 2.3045 \\
& CDL2 $\downarrow$ & 0.0121 & 0.0145 & 0.0188 & \textbf{0.0110} & \textbf{0.0110} & \textbf{0.0110} & 0.0114 & 0.0134\\
& EMD $\downarrow$ & 3.4081 & 3.4962 & 3.7561 & 3.3425 & \underline{3.3388} & \textbf{3.3364} & 3.3907 & 3.6001 \\
& F-score $\uparrow$ & 0.4654 & 0.4387 & 0.3843 & 0.4866 & \underline{0.4869} & \textbf{0.4889} & 0.4747 & 0.4292\\

\midrule

\multirow{5}{*}{Objaverse}
& Num $\downarrow$ & 731{,}072 & 500{,}000 & 624{,}640 & 787{,}174 & 710{,}859 & 762{,}776 & 358{,}357 & 106{,}814 \\
& CDL1 $\downarrow$ & 2.1798 & 2.3196 & 3.2233 & \textbf{2.0218} & \underline{2.0240} & 2.0362 & 2.1578 & 2.4546 \\
& CDL2 $\downarrow$ & 0.0122 & 0.0153 & 0.0842 & \textbf{0.0104} & \underline{0.0105} & \underline{0.0105} & 0.0124 & 0.0172 \\
& EMD $\downarrow$ & 3.3885 & 3.5664 & 4.8408 & \textbf{3.1973} & \underline{3.2025} & 3.2096 & 3.4889 & 4.1371 \\
& F-score $\uparrow$ & 0.4866 & 0.4442 & 0.3679 & \underline{0.5357} & \textbf{0.5357} & 0.5311 & 0.5059 & 0.4330\\

\midrule

\multirow{5}{*}{Wild}
& Num $\downarrow$ & 731{,}072 & 500{,}000 & 624{,}640 & 416{,}054 & 380{,}769 & 403{,}306 & 234{,}999 & 72{,}862 \\
& CDL1 $\downarrow$ & 1.7899 & 1.9798 & 2.3475 & \underline{1.4525} & \textbf{1.4514} & 1.5090 & 1.6680 & 1.8972\\
& CDL2 $\downarrow$ & 0.0089 & 0.0114 & 0.0184 & \underline{0.0055} & \textbf{0.0055} & 0.0058 & 0.0072 & 0.0098 \\
& EMD $\downarrow$ & 2.9302 & 3.2289 & 3.8451 & \underline{2.4976} & \textbf{2.4949} & 2.5602 & 2.8753 & 3.4419 \\
& F-score $\uparrow$ & 0.6324 & 0.5766 & 0.4895 & \textbf{0.7552} & \underline{0.7546} & 0.7370 & 0.6827 & 0.5997 \\

\bottomrule
\end{tabular}}
\end{table*}

\begin{figure*}[t]
    \centering
    \includegraphics[width=1\linewidth]{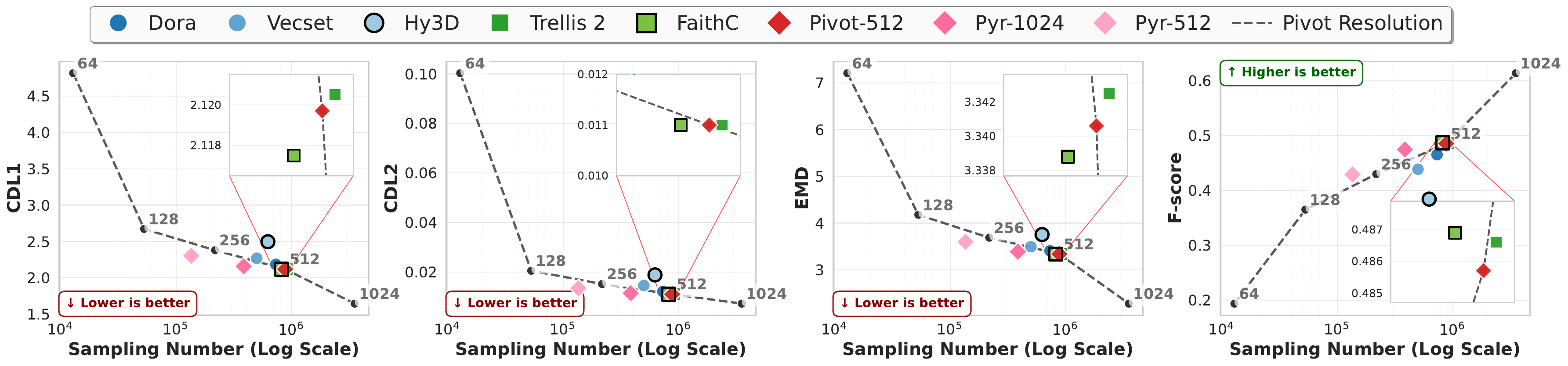}
    \caption{
    \textbf{Efficiency vs. quality trade-off for Table~\ref{tab:metrics_512}.}
    We report Chamfer Distance ($L_1/L_2$), EMD, and F-score on ABO, using the number of sampling primitives in log scale as a proxy for storage and computational cost.
    Our methods (red diamonds) achieve better Pareto efficiency than SDF-based methods (blue circles) and Dual Contouring methods (green squares).
    Notably, Pivot-512 reaches reconstruction quality comparable to FaithC with fewer active voxels.}
    \label{fig:quan_fig}
\end{figure*}

\subsection{Main Results}
\mypara{Quantitative Evaluation of P2Voxel}
Table~\ref{tab:metrics_512} summarizes reconstruction quality at 512 resolution on ABO, Objaverse, and Wild using CDL1/CDL2, EMD, and F-score ($\tau=0.002$). Overall, our \textbf{P2Voxel} (Dim~4) achieves a strong efficiency-fidelity balance, outperforming SDF-based baselines while remaining competitive with dual contouring methods that use substantially larger voxel dimensions (e.g., FaithC with Dim~18). On \textbf{ABO}, \textbf{Pivot-512} provides the best overall reconstruction quality among our variants, and \textbf{Pyramid-1024} reduces the sampling count by over $2\times$ (385k vs.\ 880k) with only a small reduction in accuracy. On \textbf{Objaverse} and \textbf{Wild}, our method maintains stable performance across metrics, while \textbf{Pyramid-512} achieves the most aggressive compression (around $\mathbf{6.5\times}$ fewer samples) at the expense of reduced reconstruction fidelity, highlighting the importance of hierarchical lifting for preserving fine details.
Figure~\ref{fig:quan_fig} analyzes the efficiency-quality trade-off on the ABO dataset. The $x$-axis denotes the number of sampling primitives (log scale), serving as a proxy for storage and computation costs. Overall, our methods (red diamonds) consistently achieve better Pareto efficiency than SDF-based sampling baselines (blue circles) and dual contouring methods (green squares). In particular, \textbf{Pivot-512} outperforms naïve resolution scaling (dashed line), achieving high reconstruction fidelity while using fewer active voxels.

\begin{table*}[t]
\centering
\caption{\textbf{Quantitative evaluation of geometry reconstruction quality of PyramidVAE.}
We report Chamfer Distance (CDL1/CDL2), Earth Mover's Distance (EMD), and F-score ($\tau=0.002$) on ABO, Objaverse, and Wild datasets at \textbf{512} resolution.
\textbf{Bold} indicates the best performance, and \underline{underlined} denotes the second best. Qualitative results are provided in~\textbf{Figure~\ref{fig:quali_vae} of Appendix~\ref{app:qualitative}}.}
\label{tab:vae_512}
\setlength{\tabcolsep}{5pt}
\resizebox{0.9\textwidth}{!}{
\begin{tabular}{l l c c c c c c}
\toprule
\multirow{2}{*}{Dataset}
 & \multirow{2}{*}{Metrics}
 & \multicolumn{3}{c}{SDF} 
 & \multicolumn{1}{c}{DC} 
 & \multicolumn{2}{c}{Ours} \\
\cmidrule(lr){3-5} \cmidrule(lr){6-6} \cmidrule(lr){7-8}
 & & Dora & Vecset & Hunyuan3D & Trellis 2 & Pivot-512 & Pyramid-512 \\
\midrule

\multirow{4}{*}{ABO}
& CDL1 $\downarrow$ & 2.2585 & 3.4681 & 2.5242 & \textbf{2.1741} & \underline{2.1908} & 2.7853 \\
& CDL2 $\downarrow$ & \underline{0.0138} & 0.0404 & 0.0355 & \textbf{0.0116} & \textbf{0.0116} & 0.0231 \\
& EMD $\downarrow$ & 3.4906 & 4.8132 & 3.9635 & \textbf{3.3969} & \underline{3.4202} & 4.2214 \\
& F-score $\uparrow$ & 0.4475 & 0.3125 & 0.4108 & \textbf{0.4725} & \underline{0.4606} & 0.3594 \\

\midrule

\multirow{4}{*}{Objaverse}
& CDL1 $\downarrow$ & 2.3785 & 3.0965 & 2.8737 & \underline{2.1071} & \textbf{2.1019} & 2.6990 \\
& CDL2 $\downarrow$ & 0.0159 & 0.0299 & 0.0671 & \underline{0.0114} & \textbf{0.0112} & 0.0218 \\
& EMD $\downarrow$ & 3.6592 & 4.4018 & 4.6346 & \textbf{3.2889} & \underline{3.2926} & 4.3712 \\
& F-score $\uparrow$ & 0.4487 & 0.3345 & 0.4028 & \textbf{0.5134} & \underline{0.5077} & 0.3951 \\

\midrule

\multirow{4}{*}{Wild}
& CDL1 $\downarrow$ & 1.9505 & 3.0526 & 2.2502 & \underline{1.7363} & \textbf{1.6202} & 2.3525 \\
& CDL2 $\downarrow$ & 0.0116 & 0.0280 & 0.0233 & \underline{0.0070} & \textbf{0.0068} & 0.0175 \\
& EMD $\downarrow$ & 3.1651 & 4.1121 & 3.7488 & \underline{2.8409} & \textbf{2.6872} & 4.0763 \\
& F-score $\uparrow$ & 0.5879 & 0.2427 & 0.5238 & \textbf{0.7100} & \underline{0.6885} & 0.4968 \\

\bottomrule
\end{tabular}}
\end{table*}

\mypara{Quantitative Evaluation of PyramidVAE}
Table~\ref{tab:vae_512} shows that \textbf{Pivot-512} achieves the best or second-best results on most metrics across ABO, Objaverse, and Wild, demonstrating the effectiveness of compact Dim~4 pivot tokens for preserving local surface evidence. 
Compared with SDF-based baselines, it consistently improves reconstruction accuracy and remains competitive with the DC-based TRELLIS~2, even outperforming it on Objaverse CDL1/CDL2 and Wild CDL1/CDL2/EMD. 
It is worth noting that the Pyramid VAE is trained only on the ABO training split with limited data, without using Objaverse or Wild for training. 
Under this setting, the VAE-compressed \textbf{Pyramid-512} inevitably sacrifices some reconstruction fidelity compared with direct Pivot-512, but still provides a compact multi-resolution latent representation with cross-dataset generalization potential for downstream token-based learning and generation.

\begin{table*}[t]
\centering
\caption{\textbf{Ablation study on pivot sampling resolution} (ranging from 64 to 1024) using the ABO dataset. 
TopoC and HighCurv represent the percentage of samples suffering from topological complexity errors and high-curvature artifacts, respectively. 
\textit{All} indicates the overall percentage of samples exhibiting \textbf{any} of the aforementioned defects. 
The results demonstrate that increasing the pivot sampling resolution mitigates both types of geometric artifacts.}
\label{tab:abo_sampling_ablation}
\setlength{\tabcolsep}{5pt}
\resizebox{\textwidth}{!}{
\begin{tabular}{c r c c c c c c c}
\toprule
Resolution & Num & CDL1$\downarrow$ & CDL2$\downarrow$ & EMD$\downarrow$ & F-score$\uparrow$ & TopoC (\%) & HighCurv (\%) & All (\%) \\
\midrule
64   &   12{,}856   & 4.8159  & 0.1003 & 7.2071 & 0.1937 & 14.85 & 17.39 & 19.09 \\
128  &   52{,}944   & 2.6730  & 0.0205 & 4.1791 & 0.3651 &  7.82 & 10.17 & 10.69 \\
256  &  217{,}834   & 2.3806  & 0.0151 & 3.6894 & 0.4299 &  3.42 &  5.27 &  5.50 \\
512  &  880{,}134   & 2.1166 & 0.0110 & 3.3364 & 0.4889 &  0.63 &  1.86 &  1.89 \\
1024 & 3{,}524{,}752 & 1.6463  & 0.0072 & 2.2793 & 0.6139 &  0.22 &  0.87 &  0.89 \\
\bottomrule
\end{tabular}}
\end{table*}

\subsection{Ablation Study and Analysis}

\mypara{Local Plannarity Analysis}
As provided in \textbf{Appendix~\ref{app:pivot_feas}}, Figure~\ref{fig:artifact} visualizes the artifact ratio distribution from Table~\ref{tab:abo_sampling_ablation}, including TopoC, HighCurv, and their union (\textit{All}). As the resolution increases from 64 to 1024, all three ratios consistently decrease, indicating that higher-resolution pivot sampling effectively mitigates both topological complexity and high-curvature defects. Notably, the reduction becomes less pronounced beyond 512, suggesting that \textbf{512} offers a practical trade-off between artifact suppression and sampling cost.

\mypara{Resolution Evaluation}
Table~\ref{tab:abo_sampling_ablation} presents an ablation study on pivot sampling resolution (64$\rightarrow$1024) on ABO. Increasing resolution consistently improves reconstruction quality: CDL1/CDL2 decrease from 4.8159/0.1003 to 1.6463/0.0072, EMD drops from 7.2071 to 2.2793, and F-score rises from 0.1937 to 0.6139. Meanwhile, artifact rates are greatly reduced, with TopoC and HighCurv decreasing from 14.85\%/17.39\% to 0.22\%/0.87\%, and overall defects (\textit{All}) dropping from 19.09\% to 0.89\%. 
Notably, the marginal gains diminish at higher resolutions (e.g., 512$\rightarrow$1024), making \textbf{512} a practical trade-off between reconstruction fidelity and sampling cost.

\begin{wraptable}{r}{0.52\textwidth}
\vspace{-8pt}
\centering
\caption{Ablation study on sampling cutting ratios at 512 resolution. 
We evaluate four configurations over levels $\{128,256,512\}$: 
S1$=[0.85,0.12,0.03]$, S2$=[0.85,0.10,0.05]$, 
S3$=[0.80,0.10,0.10]$, and S4$=[0.70,0.20,0.10]$. 
Lower thresholds activate more voxels and improve reconstruction quality.}
\label{tab:q_cuts_ablation}
\setlength{\tabcolsep}{3pt}
\renewcommand{\arraystretch}{0.95}
\resizebox{0.52\textwidth}{!}{
\begin{tabular}{l c c c c c}
\toprule
Config. & Num & CDL1$\downarrow$ & CDL2$\downarrow$ & EMD$\downarrow$ & F-score$\uparrow$ \\
\midrule
S1 & 116,992 & 2.3154 & 0.0134 & 3.6121 & 0.4234 \\
S2 & 135,736 & 2.3045 & 0.0134 & 3.6001 & 0.4292 \\
S3 & 189,783 & 2.2533 & 0.0126 & 3.5270 & 0.4437 \\
S4 & 207,885 & 2.2229 & 0.0122 & 3.4801 & 0.4547 \\
\bottomrule
\end{tabular}
}
\end{wraptable}
\textbf{Allocation Evaluation.}
Table~\ref{tab:q_cuts_ablation} quantifies how the sampling-map cut ratios control the sampling budget at maximum resolution ($R{=}512$) under four threshold settings (S1--S4). 
Relaxing the thresholds activates more fine voxels (116,992 $\rightarrow$ 207,885), yielding consistent quality gains (CDL1: 2.3154 $\rightarrow$ 2.2229; F-score: 0.4234 $\rightarrow$ 0.4547). 
Figure~\ref{fig:block} further visualizes the induced block-level distribution at $R{=}512$: increasing the budget converts more fine blocks from inactive to active and raises per-block sample counts, while most blocks remain sparse. 
This suggests that additional primitives are selectively allocated to geometrically demanding regions, such as thin structures and high-curvature boundaries, rather than uniformly inflating all areas. 
Overall, the sampling map provides a simple control knob to trade efficiency for fidelity by modulating the number of active voxels and sampled primitives under uniform cut-ratio settings. Statistics are provided in ~\textbf{Figure~\ref{fig:block} of Appendix~\ref{app:q_cuts_vis}}.

\section{Conclusion}

We propose \textbf{P2Voxel}, a compact mesh tokenization framework that reformulates voxel-based mesh representation as \emph{local surface evidence sampling}. 
By encoding each active voxel with a pivot point and an orientation sign, P2Voxel replaces dense SDF storage with a minimal Dim~4 descriptor that can induce the corner values required for deterministic Sparse Marching Cubes reconstruction. 
Built on local planarity and spatial complexity assumptions, the pyramid pivot representation adaptively allocates finer tokens to geometrically complex regions while keeping smooth regions coarse, achieving a better balance between reconstruction fidelity and token efficiency.
Experiments on ABO, Objaverse, and Wild show that P2Voxel provides competitive reconstruction quality with substantially fewer primitives than representative SDF- and contouring-based baselines. 
By turning dense mesh geometry into compact, locally reconstructable pyramid pivot blocks, P2Voxel provides a concrete token space for learning-based 3D representation. 
Future work will further explore its use in latent modeling and diffusion-based shape generation.

\bibliographystyle{ieeetr}
\bibliography{reference}

@String { ICCV     = {ICCV} }

@String { TOG      = {ACM Trans. Graph.} }

@String{Computer = "{IEEE} Computer" }

@String{Springer = "Springer-Verlag" }

@article{luo2025faithful,
  title={Faithful Contouring: Near-Lossless 3D Voxel Representation Free from Iso-surface},
  author={Luo, Yihao and He, Xianglong and Pan, Chuanyu and Chen, Yiwen and Wu, Jiaqi and Li, Yangguang and Ouyang, Wanli and Hu, Yuanming and Yang, Guang and Yap, ChoonHwai},
  journal={arXiv preprint arXiv:2511.04029},
  year={2025}
}

@article{zhang20233dshape2vecset,
  title={3dshape2vecset: A 3d shape representation for neural fields and generative diffusion models},
  author={Zhang, Biao and Tang, Jiapeng and Niessner, Matthias and Wonka, Peter},
  journal={ACM Transactions On Graphics (TOG)},
  volume={42},
  number={4},
  pages={1--16},
  year={2023},
  publisher={ACM New York, NY, USA}
}

@article{zhao2024flexidreamerflexicubes,
  title={Flexidreamer: single image-to-3d generation with flexicubes},
  author={Zhao, Ruowen and Wang, Zhengyi and Wang, Yikai and Zhou, Zihan and Zhu, Jun},
  journal={arXiv preprint arXiv:2404.00987},
  year={2024}
}

@inproceedings{xiang2025structuredtrellis1,
  title={Structured 3d latents for scalable and versatile 3d generation},
  author={Xiang, Jianfeng and Lv, Zelong and Xu, Sicheng and Deng, Yu and Wang, Ruicheng and Zhang, Bowen and Chen, Dong and Tong, Xin and Yang, Jiaolong},
  booktitle={Proceedings of the Computer Vision and Pattern Recognition Conference},
  pages={21469--21480},
  year={2025}
}

@article{xiang2025nativetrellis2,
  title={Native and Compact Structured Latents for 3D Generation},
  author={Xiang, Jianfeng and Chen, Xiaoxue and Xu, Sicheng and Wang, Ruicheng and Lv, Zelong and Deng, Yu and Zhu, Hongyuan and Dong, Yue and Zhao, Hao and Yuan, Nicholas Jing and others},
  journal={arXiv preprint arXiv:2512.14692},
  year={2025}
}

@article{nichol2022pointe,
  title={Point-e: A system for generating 3d point clouds from complex prompts},
  author={Nichol, Alex and Jun, Heewoo and Dhariwal, Prafulla and Mishkin, Pamela and Chen, Mark},
  journal={arXiv preprint arXiv:2212.08751},
  year={2022}
}

@article{jun2023shap-e,
  title={Shap-e: Generating conditional 3d implicit functions},
  author={Jun, Heewoo and Nichol, Alex},
  journal={arXiv preprint arXiv:2305.02463},
  year={2023}
}

@inproceedings{nash2020polygen,
  title={Polygen: An autoregressive generative model of 3d meshes},
  author={Nash, Charlie and Ganin, Yaroslav and Eslami, SM Ali and Battaglia, Peter},
  booktitle={International conference on machine learning},
  pages={7220--7229},
  year={2020},
  organization={PMLR}
}

@inproceedings{siddiqui2024meshgpt,
  title={Meshgpt: Generating triangle meshes with decoder-only transformers},
  author={Siddiqui, Yawar and Alliegro, Antonio and Artemov, Alexey and Tommasi, Tatiana and Sirigatti, Daniele and Rosov, Vladislav and Dai, Angela and Nie{\ss}ner, Matthias},
  booktitle={Proceedings of the IEEE/CVF conference on computer vision and pattern recognition},
  pages={19615--19625},
  year={2024}
}

@article{mildenhall2021nerf,
  title={Nerf: Representing scenes as neural radiance fields for view synthesis},
  author={Mildenhall, Ben and Srinivasan, Pratul P and Tancik, Matthew and Barron, Jonathan T and Ramamoorthi, Ravi and Ng, Ren},
  journal={Communications of the ACM},
  volume={65},
  number={1},
  pages={99--106},
  year={2021},
  publisher={ACM New York, NY, USA}
}

@inproceedings{oleynikova2016signedsdf,
  title={Signed distance fields: A natural representation for both mapping and planning},
  author={Oleynikova, Helen and Millane, Alexander and Taylor, Zachary and Galceran, Enric and Nieto, Juan and Siegwart, Roland},
  booktitle={RSS 2016 workshop: geometry and beyond-representations, physics, and scene understanding for robotics},
  year={2016},
  organization={University of Michigan}
}

@inproceedings{chen2025dora,
  title={Dora: Sampling and benchmarking for 3d shape variational auto-encoders},
  author={Chen, Rui and Zhang, Jianfeng and Liang, Yixun and Luo, Guan and Li, Weiyu and Liu, Jiarui and Li, Xiu and Long, Xiaoxiao and Feng, Jiashi and Tan, Ping},
  booktitle={Proceedings of the Computer Vision and Pattern Recognition Conference},
  pages={16251--16261},
  year={2025}
}

@article{shen2021deepDMTet,
  title={Deep marching tetrahedra: a hybrid representation for high-resolution 3d shape synthesis},
  author={Shen, Tianchang and Gao, Jun and Yin, Kangxue and Liu, Ming-Yu and Fidler, Sanja},
  journal={Advances in Neural Information Processing Systems},
  volume={34},
  pages={6087--6101},
  year={2021}
}

@inproceedings{deng2025efficientoctrees,
  title={Efficient autoregressive shape generation via octree-based adaptive tokenization},
  author={Deng, Kangle and Liu, Hsueh-Ti Derek and Zhu, Yiheng and Sun, Xiaoxia and Shang, Chong and Bhat, Kiran S and Ramanan, Deva and Zhu, Jun-Yan and Agrawala, Maneesh and Zhou, Tinghui},
  booktitle={Proceedings of the IEEE/CVF International Conference on Computer Vision},
  pages={11685--11696},
  year={2025}
}

@inproceedings{ju2002dualhashgrids,
  title={Dual contouring of hermite data},
  author={Ju, Tao and Losasso, Frank and Schaefer, Scott and Warren, Joe},
  booktitle={Proceedings of the 29th annual conference on Computer graphics and interactive techniques},
  pages={339--346},
  year={2002}
}

@inproceedings{ju2002dualDC,
  title={Dual contouring of hermite data},
  author={Ju, Tao and Losasso, Frank and Schaefer, Scott and Warren, Joe},
  booktitle={Proceedings of the 29th annual conference on Computer graphics and interactive techniques},
  pages={339--346},
  year={2002}
}

@incollection{lorensen1998marching,
  title={Marching cubes: A high resolution 3D surface construction algorithm},
  author={Lorensen, William E and Cline, Harvey E},
  booktitle={Seminal graphics: pioneering efforts that shaped the field},
  pages={347--353},
  year={1998}
}

@misc{tang2022cubvh,
  author = {Tang, Jiaxiang},
  title = {cuBVH: A CUDA Mesh BVH acceleration toolkit},
  year = {2022},
  publisher = {GitHub},
  journal = {GitHub repository},
  howpublished = {\url{https://github.com/ashawkey/cubvh}},
}

@article{zhou2018open3d,
  title={Open3D: A modern library for 3D data processing},
  author={Zhou, Qian-Yi and Park, Jaesik and Koltun, Vladlen},
  journal={arXiv preprint arXiv:1801.09847},
  year={2018}
}

@inproceedings{dyken2008paralelmc,
  title={High-speed marching cubes using histopyramids},
  author={Dyken, Christopher and Ziegler, Gernot and Theobalt, Christian and Seidel, Hans-Peter},
  booktitle={Computer Graphics Forum},
  volume={27},
  number={8},
  pages={2028--2039},
  year={2008},
  organization={Wiley Online Library}
}

@article{ozkan1994pocs,
  title={POCS-based restoration of space-varying blurred images},
  author={Ozkan, Mehmet K and Tekalp, A Murat and Sezan, M Ibrahim},
  journal={IEEE Transactions on Image Processing},
  volume={3},
  number={4},
  pages={450--454},
  year={1994},
  publisher={IEEE}
}

@article{chen2024meshanything,
  title={Meshanything: Artist-created mesh generation with autoregressive transformers},
  author={Chen, Yiwen and He, Tong and Huang, Di and Ye, Weicai and Chen, Sijin and Tang, Jiaxiang and Chen, Xin and Cai, Zhongang and Yang, Lei and Yu, Gang and others},
  journal={arXiv preprint arXiv:2406.10163},
  year={2024}
}

@InProceedings{Chen_2025_ICCV,
    author    = {Chen, Yiwen and Wang, Yikai and Luo, Yihao and Wang, Zhengyi and Chen, Zilong and Zhu, Jun and Zhang, Chi and Lin, Guosheng},
    title     = {MeshAnything V2: Artist-Created Mesh Generation with Adjacent Mesh Tokenization},
    booktitle = {Proceedings of the IEEE/CVF International Conference on Computer Vision (ICCV)},
    month     = {October},
    year      = {2025},
    pages     = {13922-13931}
}

@inproceedings{weng2025scaling,
  title={Scaling mesh generation via compressive tokenization},
  author={Weng, Haohan and Zhao, Zibo and Lei, Biwen and Yang, Xianghui and Liu, Jian and Lai, Zeqiang and Chen, Zhuo and Liu, Yuhong and Jiang, Jie and Guo, Chunchao and others},
  booktitle={Proceedings of the Computer Vision and Pattern Recognition Conference},
  pages={11093--11103},
  year={2025}
}

@book{foley1996computer,
  title={Computer graphics: principles and practice},
  author={Foley, James D},
  volume={12110},
  year={1996},
  publisher={Addison-Wesley Professional}
}

@book{shirley2009fundamentals,
  title={Fundamentals of computer graphics},
  author={Shirley, Peter and Ashikhmin, Michael and Marschner, Steve},
  year={2009},
  publisher={AK Peters/CRC Press}
}

@article{yang2021geometry,
  title={Geometry processing with neural fields},
  author={Yang, Guandao and Belongie, Serge and Hariharan, Bharath and Koltun, Vladlen},
  journal={Advances in Neural Information Processing Systems},
  volume={34},
  pages={22483--22497},
  year={2021}
}

@book{akenine2019real,
  title={Real-time rendering},
  author={Akenine-Moller, Tomas and Haines, Eric and Hoffman, Naty},
  year={2019},
  publisher={AK Peters/crc Press}
}

@inproceedings{saito1990comprehensible,
  title={Comprehensible rendering of 3-D shapes},
  author={Saito, Takafumi and Takahashi, Tokiichiro},
  booktitle={Proceedings of the 17th annual conference on Computer graphics and interactive techniques},
  pages={197--206},
  year={1990}
}

@article{zangi2004water,
  title={Water confined to a slab geometry: a review of recent computer simulation studies},
  author={Zangi, Ronen},
  journal={Journal of Physics: Condensed Matter},
  volume={16},
  number={45},
  pages={S5371},
  year={2004},
  publisher={IoP Publishing}
}

@article{gao2022nerf,
  title={Nerf: Neural radiance field in 3d vision, a comprehensive review},
  author={Gao, Kyle and Gao, Yina and He, Hongjie and Lu, Dening and Xu, Linlin and Li, Jonathan},
  journal={arXiv preprint arXiv:2210.00379},
  year={2022}
}

@article{jones20063d,
  title={3D distance fields: A survey of techniques and applications},
  author={Jones, Mark W and Baerentzen, J Andreas and Sramek, Milos},
  journal={IEEE Transactions on visualization and Computer Graphics},
  volume={12},
  number={4},
  pages={581--599},
  year={2006},
  publisher={IEEE}
}

@article{sitzmann2020implicit_siren,
  title={Implicit neural representations with periodic activation functions},
  author={Sitzmann, Vincent and Martel, Julien and Bergman, Alexander and Lindell, David and Wetzstein, Gordon},
  journal={Advances in neural information processing systems},
  volume={33},
  pages={7462--7473},
  year={2020}
}

@article{zhao2025hunyuan3d,
  title={Hunyuan3d 2.0: Scaling diffusion models for high resolution textured 3d assets generation},
  author={Zhao, Zibo and Lai, Zeqiang and Lin, Qingxiang and Zhao, Yunfei and Liu, Haolin and Yang, Shuhui and Feng, Yifei and Yang, Mingxin and Zhang, Sheng and Yang, Xianghui and others},
  journal={arXiv preprint arXiv:2501.12202},
  year={2025}
}

@inproceedings{collins2022abo,
  title={Abo: Dataset and benchmarks for real-world 3d object understanding},
  author={Collins, Jasmine and Goel, Shubham and Deng, Kenan and Luthra, Achleshwar and Xu, Leon and Gundogdu, Erhan and Zhang, Xi and Vicente, Tomas F Yago and Dideriksen, Thomas and Arora, Himanshu and others},
  booktitle={Proceedings of the IEEE/CVF conference on computer vision and pattern recognition},
  pages={21126--21136},
  year={2022}
}

@inproceedings{deitke2023objaverse,
  title={Objaverse: A universe of annotated 3d objects},
  author={Deitke, Matt and Schwenk, Dustin and Salvador, Jordi and Weihs, Luca and Michel, Oscar and VanderBilt, Eli and Schmidt, Ludwig and Ehsani, Kiana and Kembhavi, Aniruddha and Farhadi, Ali},
  booktitle={Proceedings of the IEEE/CVF conference on computer vision and pattern recognition},
  pages={13142--13153},
  year={2023}
}

@inproceedings{fan2017point_chamfer,
  title={A point set generation network for 3d object reconstruction from a single image},
  author={Fan, Haoqiang and Su, Hao and Guibas, Leonidas J},
  booktitle={Proceedings of the IEEE conference on computer vision and pattern recognition},
  pages={605--613},
  year={2017}
}

@article{rubner2000earth,
  title={The earth mover's distance as a metric for image retrieval},
  author={Rubner, Yossi and Tomasi, Carlo and Guibas, Leonidas J},
  journal={International journal of computer vision},
  volume={40},
  number={2},
  pages={99--121},
  year={2000},
  publisher={Springer}
}

@inproceedings{sturm2012benchmark_fscore,
  title={A benchmark for the evaluation of RGB-D SLAM systems},
  author={Sturm, J{\"u}rgen and Engelhard, Nikolas and Endres, Felix and Burgard, Wolfram and Cremers, Daniel},
  booktitle={2012 IEEE/RSJ international conference on intelligent robots and systems},
  pages={573--580},
  year={2012},
  organization={IEEE}
}

@inproceedings{graham20183d,
  title={3d semantic segmentation with submanifold sparse convolutional networks},
  author={Graham, Benjamin and Engelcke, Martin and Van Der Maaten, Laurens},
  booktitle={Proceedings of the IEEE conference on computer vision and pattern recognition},
  pages={9224--9232},
  year={2018}
}

@inproceedings{wu20153d,
  title={3d shapenets: A deep representation for volumetric shapes},
  author={Wu, Zhirong and Song, Shuran and Khosla, Aditya and Yu, Fisher and Zhang, Linguang and Tang, Xiaoou and Xiao, Jianxiong},
  booktitle={Proceedings of the IEEE conference on computer vision and pattern recognition},
  pages={1912--1920},
  year={2015}
}

\clearpage
\appendix

\section{Pivot Feasibility within a Voxel}
\label{app:pivot_feas}

\begin{wrapfigure}{r}{0.44\textwidth}
    \centering
    \includegraphics[width=0.99\linewidth]{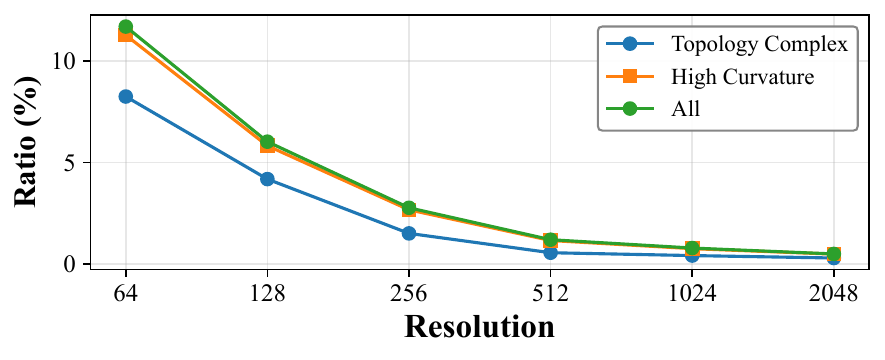}
    \caption{\textbf{Scale effect of mesh in Figure ~\ref{fig:pivot_pipe1}.} 
    The ratios of topology-complex and high-curvature regions decrease with increasing resolution, especially beyond $r\geq 1024$. 
    Please also refer to Table~\ref{tab:abo_sampling_ablation} and Figure ~\ref{fig:artifact}.}
    \label{fig:scale}
\end{wrapfigure}

\mypara{Sufficient Resolution}
Assumption~1 is only valid when the voxel size is sufficiently small such that each active voxel contains a \emph{single} and \emph{nearly planar} surface patch.
To ensure this, we perform a local complexity test by probing multiple sub-voxel locations within each candidate voxel and querying their closest surface points and corresponding triangle normals.
A voxel is labeled as \emph{ambiguous} if the normals exhibit strong directional inconsistency (suggesting multi-sheet or topologically complex regions) or if the probed points deviate noticeably from a single plane (indicating high curvature).
In our implementation, we use a normal-consistency threshold of $0.9$ (approximately a $25^\circ$ tolerance) for the topological check and a planarity tolerance of $5\%$ of the voxel size for high curvature, which provides a conservative filter for valid locally planar voxels, where Figure ~\ref{fig:scale} demonstrates the trends of scaling voxelization resolution.

\mypara{Pivot Feasibility within a Voxel}
To make the pivot token well-defined and numerically stable, the pivot point must lie strictly inside its associated voxel (otherwise, the voxel index recovered from $\mathbf{p}$ becomes ambiguous), and it should not collapse to the voxel center $\mathbf{c}$ (otherwise, the normal direction becomes ill-conditioned).
Given an initial pivot estimate on the plane, we refine it by solving a constrained closest-point problem via \emph{alternating projections} (projection onto convex sets, POCS~\cite{ozkan1994pocs}): we first project $\mathbf{c}$ onto the plane, then alternately enforce \emph{in-voxel} feasibility (by clamping to the voxel bounds with a small $\epsilon$ margin) and \emph{on-plane} feasibility (by re-projecting onto the plane) for a few iterations, and finally apply a hard in-voxel constraint to guarantee that the stored pivot remains inside the voxel.

\begin{figure*}[t]
    \centering
    \includegraphics[width=1\linewidth]{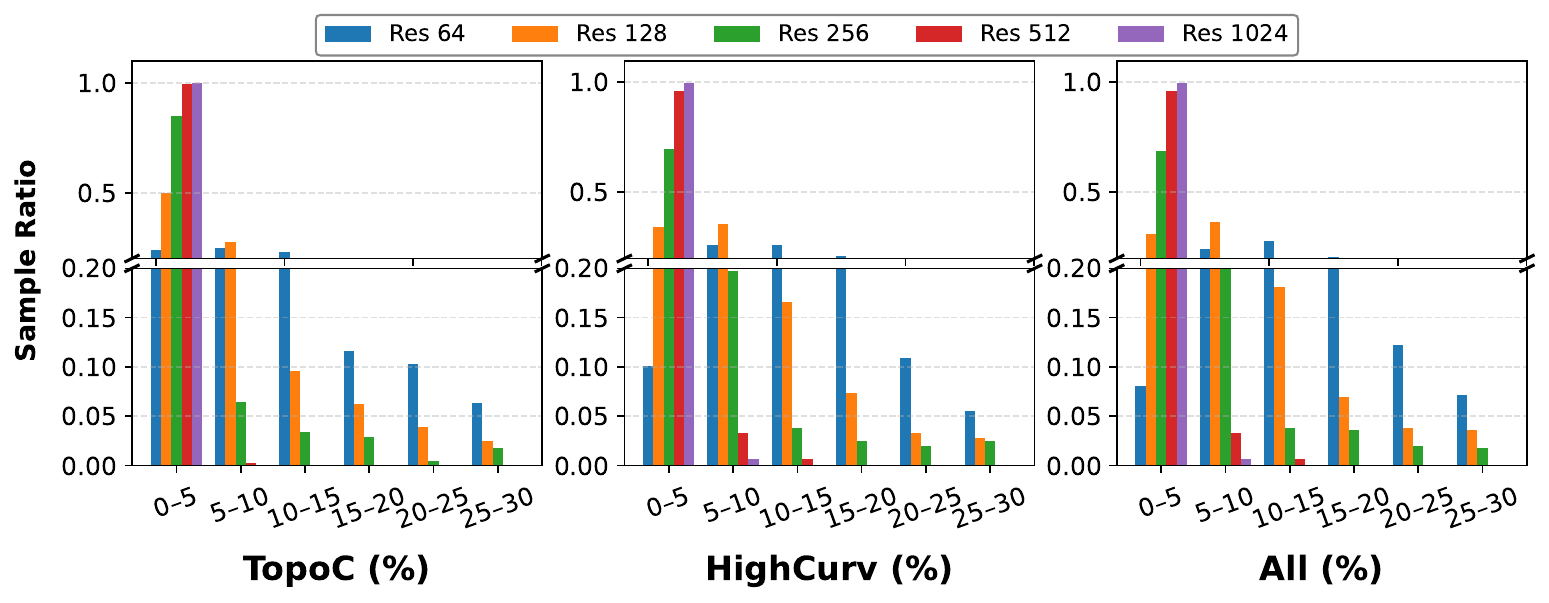}
    \caption{\textbf{Artifact ratio distribution across pivot sampling resolutions for Table~\ref{tab:abo_sampling_ablation}.} We visualize the percentage of samples affected by topological complexity (TopoC), high-curvature artifacts (HighCurv), and their union (\textit{All}) under different pivot sampling resolutions (64--1024) on ABO. Artifact ratios consistently decrease as resolution increases, while improvements beyond \textbf{512} become less pronounced, suggesting \textbf{512} as a practical trade-off between robustness and sampling cost.}
    \label{fig:artifact}
\end{figure*}

\mypara{Artifact Ratio Distribution across Resolutions}
We analyze the effect of pivot sampling resolution on reconstruction robustness using ABO. 
For resolutions from $64$ to $1024$, we report the percentage of samples affected by topological complexity (TopoC), high-curvature artifacts (HighCurv), and their union (\textit{All}). 
The artifact ratios decrease consistently as resolution increases, showing that finer pivot sampling better captures challenging local geometry. 
However, the gains become marginal beyond resolution $512$, suggesting that $512$ provides a practical balance between robustness and sampling cost. 
Accordingly, we use $512$ as the default maximum resolution in our experiments.

\section{Implementation Details}
\label{app:detail}

\mypara{Baselines}
We benchmark P2Voxel against representative baselines from two mainstream geometry discretization paradigms: SDF-based sampling methods, including 3DShape2VecSet (VecSet)~\cite{zhang20233dshape2vecset}, DORA~\cite{chen2025dora}, and Hunyuan3D-2.1~\cite{zhao2025hunyuan3d}; and Dual Contouring-based methods, including FaithC~\cite{luo2025faithful} and TRELLIS~2~\cite{xiang2025nativetrellis2}. 
Our evaluation focuses on the geometry discretization stage, namely the sampling strategy and its resulting surface reconstruction fidelity, while excluding downstream generative components such as VAE tokenization, latent modeling, or diffusion synthesis. 
For a controlled comparison, all SDF-based baselines sample or downsample points from the same watertight input mesh and reconstruct surfaces using the same neural implicit representation~\cite{sitzmann2020implicit_siren}, so that performance differences mainly reflect the effect of the sampling strategy rather than reconstruction-network variations.

\mypara{Datasets and Metrics}
We conduct experiments on three disjoint test sets, each containing approximately 400 shapes. 
The \textbf{ABO} dataset~\cite{collins2022abo} contains product models from a catalog of over 147k items and 7,953 artist-designed meshes. 
The \textbf{Objaverse} dataset~\cite{deitke2023objaverse} is sampled from a large-scale repository of over 800k 3D assets, covering diverse categories such as animals, humans, vehicles, and everyday objects. 
The \textbf{Wild} dataset includes challenging internet-collected shapes with noisy geometry, irregular sampling artifacts, and complex topology.
All shapes are converted into watertight reference surfaces using DORA's UDF-to-SDF pipeline, followed by Marching Cubes and largest connected component extraction. 
This provides consistent ground-truth surfaces for evaluation while reducing ambiguity from disconnected fragments and non-manifold artifacts.
We evaluate reconstruction quality using complementary geometric metrics. 
Chamfer Distance under L1 and L2 norms (CDL1 and CDL2)~\cite{fan2017point_chamfer} measures bidirectional nearest-neighbor discrepancy between reconstructed and reference point sets. 
Earth Mover's Distance (EMD)~\cite{rubner2000earth} captures global distribution mismatch and structural distortion. 
We also report F-score~\cite{sturm2012benchmark_fscore} at a threshold of $0.002$, which balances precision and recall to indicate reconstruction completeness and surface correctness.

\begin{table*}[t]
\centering
\caption{Average runtime breakdown of pyramid voxelization at resolutions 512 and 1024. 
Values are averaged across ABO, Objaverse, and Wild datasets. 
Note that the reported times measure the end-to-end execution in \textbf{Python}, explicitly including the overall I/O (read/write) overhead.}
\label{tab:resolution_breakdown}
\setlength{\tabcolsep}{4pt}
\begin{tabular}{c c c c c}
\toprule
Resolution & \makecell{Sampling \\ map} & \makecell{Pyramid \\ Voxelization} & \makecell{Resolution \\ Lifting} & \makecell{Sparse MC \\ RTX 4090} \\
\midrule
512  & 8.39 s & 5.96 s  & 0.01 s & 0.60 s \\
1024 & 8.98 s & 14.55 s & 0.08 s & 2.34 s \\
\bottomrule
\end{tabular}
\end{table*}

\begin{table*}[t]
\centering
\caption{Runtime benchmark for training and inference.
Training speed is measured under $r_7=400{,}000$ and reported as wall-clock time per 1,000 steps, estimated total training time, and peak VRAM.
Inference speed is reported as end-to-end reconstruction time per shape and peak VRAM.}
\label{tab:runtime_benchmark}
\setlength{\tabcolsep}{6pt}
\resizebox{\textwidth}{!}{
\begin{tabular}{l l c c c c}
\toprule
Setting & GPU & Block Batch Size & Measured Runtime & Estimated Total Time & Peak VRAM \\
\midrule
\multirow{1}{*}{Train}
& RTX 4090 & 32 & 4m36s / 1,000 steps & 12.8 days & 12,299 MiB \\
\midrule
\multirow{1}{*}{Inference}
& RTX 4090 & -- & 1.69 s / shape & -- & 4,119 MiB \\
\bottomrule
\end{tabular}}
\end{table*}

\mypara{Setups}
We evaluate two voxelization configurations to study the trade-off between reconstruction fidelity and token efficiency: a single-resolution baseline \textbf{Pivot-512} and our hierarchical \textbf{Pyramid-R} strategy with $R_{\max}=32$.
In \textbf{Pivot-512}, we perform pivot-based voxelization at a fixed resolution of $512^3$, serving as a strong single-scale reference.
For the hierarchical setting, we instantiate two variants that progressively lift coarse pivot tokens to finer resolutions.
\textbf{Pyramid-512} adopts a 3-level hierarchy $\{128,256,512\}$ with allocation ratios $[0.85,0.10,0.05]$, prioritizing coarse coverage while reserving fine samples for structurally informative regions.
\textbf{Pyramid-1024} extends this to a 4-level hierarchy $\{128,256,512,1024\}$ with allocation ratios $[0.50,0.35,0.12,0.03]$, enabling higher-resolution reconstruction with limited fine-scale sampling and sufficient mid-scale support.
Unless otherwise specified, we keep all scoring hyperparameters fixed across datasets and baselines, setting $\lambda_{\mathrm{var}}=0.5$, $\lambda_{\mathrm{grad}}=0.3$, and $n=3.0$, which control curvature variance, curvature gradient magnitude, and normal deviation in the block-level importance score, respectively.
All experiments are conducted on a workstation equipped with an AMD EPYC 7543 32-Core Processor and eight NVIDIA RTX 4090 GPUs with 24GB VRAM, under identical hardware and implementation settings for fair comparison.

\mypara{Runtime Breakdown}
Table~\ref{tab:resolution_breakdown} summarizes the average end-to-end runtime of our pyramid voxelization pipeline at resolutions 512 and 1024, decomposed into four stages. The \textit{sampling map} construction dominates the preprocessing cost and remains relatively stable across resolutions (8.39s$\rightarrow$8.98s), while \textit{pyramid voxelization} increases notably with resolution (5.96s$\rightarrow$14.55s) due to the higher-density token generation and accumulation. In contrast, \textit{resolution lifting} is lightweight in both settings (0.01s and 0.08s), indicating that multi-scale refinement introduces negligible overhead compared to voxelization itself. Finally, the \textit{sparse Marching Cubes} stage scales with output resolution (0.60s$\rightarrow$2.34s) and is the only component accelerated by the \textbf{CuBVH} CUDA implementation~\cite{tang2022cubvh} on RTX 4090; all other stages are executed in \textbf{Python} and include I/O overhead. This suggests substantial room for further optimization by migrating the sampling and voxelization stages to GPU kernels and reducing Python-side bottlenecks.

\section{More Experimental Results}
\label{app:more_exp}

\begin{figure*}[t]
    \centering
    \includegraphics[width=1\linewidth]{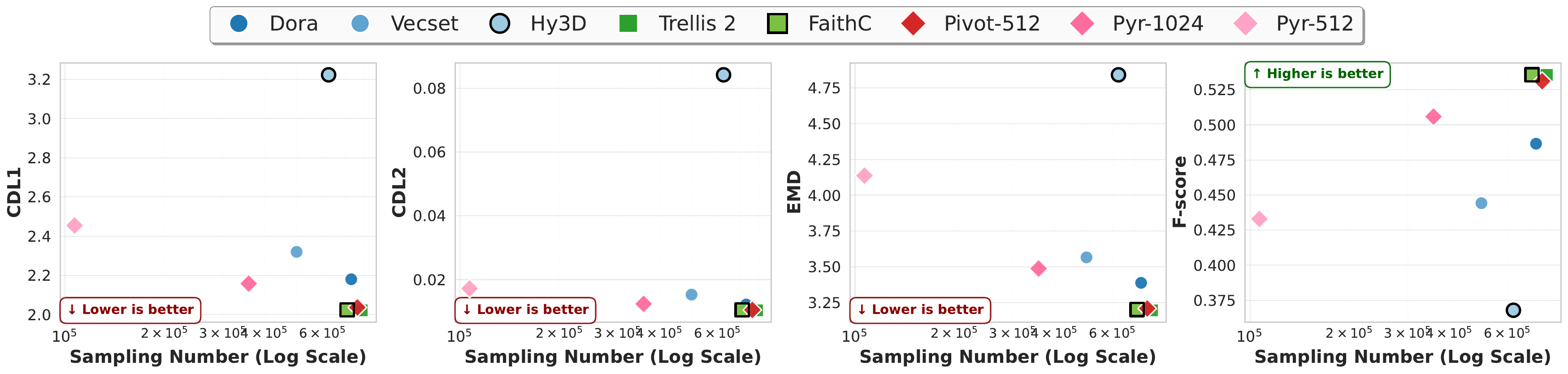}
    \caption{\textbf{Efficiency vs. quality trade-off analysis for Table~\ref{tab:metrics_512}.} We report Chamfer Distance ($L_1/L_2$), Earth Mover's Distance (EMD), and F-score on \textbf{Objaverse datasets}. The $x$-axis represents the number of sampling primitives (log scale), serving as a proxy for storage and computational cost. Red diamonds (Ours) demonstrate superior Pareto efficiency compared to SDF-based methods (blue circles) and Dual Contouring methods (green squares). }
    \label{fig:quan_fig2}
\end{figure*}

\begin{figure*}[t]
    \centering
    \includegraphics[width=1\linewidth]{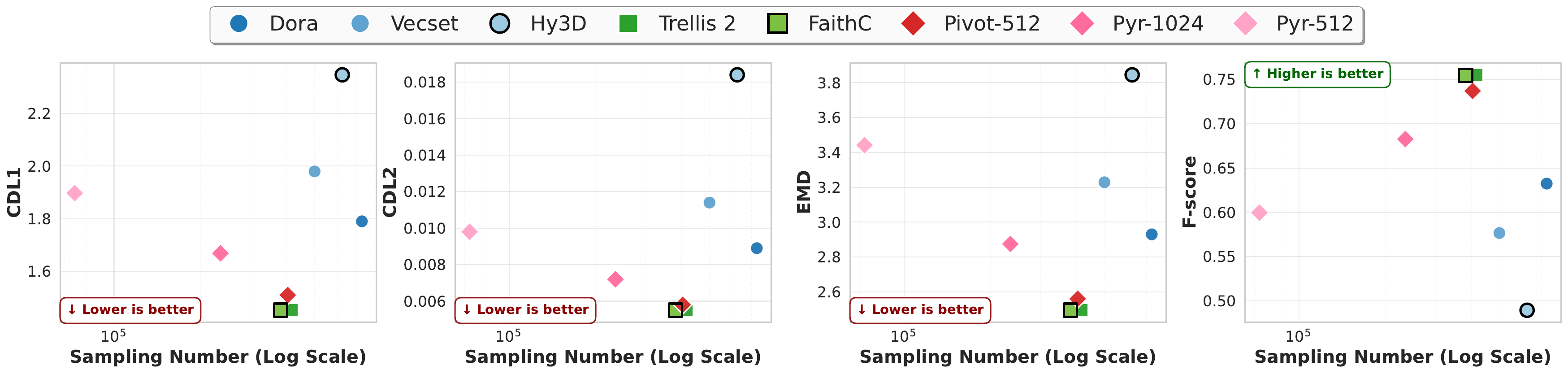}
    \caption{\textbf{Efficiency vs. quality trade-off analysis  for Table~\ref{tab:metrics_512}.} We report Chamfer Distance ($L_1/L_2$), Earth Mover's Distance (EMD), and F-score on \textbf{Wild datasets}. The $x$-axis represents the number of sampling primitives (log scale), serving as a proxy for storage and computational cost. Red diamonds (Ours) demonstrate superior Pareto efficiency compared to SDF-based methods (blue circles) and Dual Contouring methods (green squares). }
    \label{fig:quan_fig3}
\end{figure*}

\subsection{More Visualization of Efficiency-quality for Table~\ref{tab:metrics_512}}
\label{app:obj_wild}

Figs.~\ref{fig:quan_fig2} and~\ref{fig:quan_fig3} analyze the efficiency-quality trade-off using CDL1/CDL2, EMD, and F-score, where the $x$-axis denotes the number of sampling primitives (log scale) as a proxy for storage and computation costs. Across \textbf{ABO}, \textbf{Objaverse}, and \textbf{Wild}, our methods (red diamonds) consistently exhibit strong Pareto efficiency compared to SDF-based sampling baselines (blue circles) and dual contouring methods (green squares). In particular, \textbf{Pivot-512} achieves competitive reconstruction fidelity while requiring substantially fewer active voxels, demonstrating an effective balance between sampling budget and surface quality.

\begin{figure*}
    \centering
    \includegraphics[width=1\linewidth]{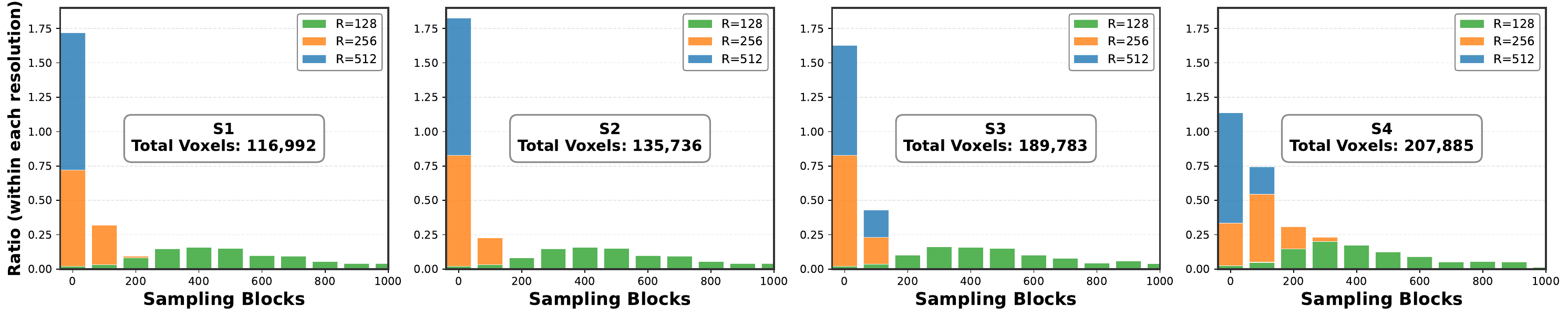}
    \caption{Visualization of sampling-map induced block activation (focus on R=512) for \textbf{Table~\ref{tab:q_cuts_ablation}}. We visualize how different sampling cut ratios redistribute the sampling budget across blocks. At the maximum resolution (R=512), increasing the total number of sampled primitives consistently activates more fine blocks and increases their per-block sample counts, resulting in clear qualitative improvements in reconstruction. This supports adaptive, region-dependent sampling allocation.}
    \label{fig:block}
\end{figure*}

\subsection{Block Activation Distribution of Table~\ref{tab:q_cuts_ablation}.}
\label{app:q_cuts_vis}

Figure~\ref{fig:block} visualizes how the sampling map redistributes the sampling budget across pyramid blocks at the maximum resolution ($R{=}512$) \textbf{Table~\ref{tab:q_cuts_ablation}}. 
As the cut ratio is relaxed, more fine-resolution blocks become active, and the number of sampled primitives within active blocks also increases. 
Importantly, the distribution remains highly sparse: most blocks still contain only a small number of samples, while additional budget is concentrated in selected regions. 
This indicates that the sampling map does not uniformly densify the entire shape, but instead expands fine-level coverage in geometrically demanding areas, showing the effectiveness of region-dependent sampling allocation.

\begin{table*}[t]
\centering
\caption{\textbf{Quantitative evaluation of geometry reconstruction quality.}
We report Chamfer Distance (CDL1/CDL2), Earth Mover's Distance (EMD), and F-score ($\tau=0.002$) on ABO, Objaverse, and Wild datasets at watertight \textbf{1024} resolution. We evaluate our \textbf{Pivot} (single-scale $512$) configurations.
Considering the trade-off between voxel dimension capacity and reconstruction fidelity, our P2Voxel method (Dim 4) achieves the most efficient performance.
Results are sorted by method category. \textbf{Bold} indicates the best performance, and \underline{underlined} denotes the second best.}
\label{tab:chamfer_1024}
\setlength{\tabcolsep}{5pt}
\begin{tabular}{l|l|ccccc|c}
\toprule
 &   Dataset     & Dora   & Vecset & Faithc & TRELLIS.2 & Hy3D & Ours-Pivot \\
\midrule
\multirow{3}{*}{CDL1 \textdownarrow}
& ABO       & 2.3144 & 2.4034 & 2.1679 & 2.1606 & 2.3303 & 2.1440 \\
& Objaverse & 2.3599 & 2.5113 & 2.0615 & 2.0534 & 2.4324 & 2.0091 \\
& Wild      & 1.9001 & 2.1017 & 1.6637 & 1.6515 & 2.0115 & 1.5304 \\
\midrule
\multirow{3}{*}{CDL2 \textdownarrow}
& ABO       & 0.0139 & 0.0167 & 0.0115 & 0.0116 & 0.0151 & 0.0113 \\
& Objaverse & 0.0162 & 0.0203 & 0.0108 & 0.0107 & 0.0180 & 0.0103 \\
& Wild     & 0.0103 & 0.0132 & 0.0073 & 0.0072 & 0.0117 & 0.0061 \\
\midrule
\multirow{3}{*}{EMD \textdownarrow}
& ABO       & 3.5541 & 3.6460 & 3.3838 & 3.3765 & 3.5992 & 3.3589 \\
& Objaverse  & 3.7049 & 3.8994 & 3.2250 & 3.2121 & 3.7182 & 3.1674 \\
& Wild     & 3.3108 & 3.6483 & 2.7409 & 2.7314 & 3.4402 & 2.5593\\
\midrule
\multirow{3}{*}{F-score \textuparrow}
& ABO       & 0.4338 & 0.4089 & 0.4713 & 0.4759 & 0.4211 & 0.4796 \\
& Objaverse & 0.4514 & 0.4121 & 0.5171 & 0.5199 & 0.4249 & 0.5345 \\
& Wild1     & 0.5961 & 0.5435 & 0.6581 & 0.6631 & 0.5760 & 0.7171 \\
\bottomrule
\end{tabular}
\end{table*}

\subsection{Higher Watertight Resolution}
\label{app:higherwater}

Table~\ref{tab:chamfer_1024} reports reconstruction results on ABO, Objaverse, and Wild when the watertight ground-truth surfaces are generated at a higher resolution of \textbf{1024}. Compared with our main results under the default watertight setting, we observe that increasing the watertight resolution leads to only minor numerical variations across all metrics, indicating that our evaluation is not sensitive to the specific watertight extraction resolution. Under this stricter setting, our \textbf{Pivot-1024} (Dim~4) remains consistently competitive, achieving strong CDL1/CDL2 and EMD with the highest or near-highest F-score across datasets. Overall, these results suggest that the reported performance trends are stable and that further increasing the watertight resolution has limited impact on the comparative conclusions.

\section{Details of Pyramid VAE}
\label{app:vae}
\paragraph{Pyramid representation and architecture.}
We train the Pyramid VAE on the ABO training split using pyramid pivot tokens extracted from watertight meshes voxelized at a global resolution of $512^3$. 
Each local block has a maximum resolution of $R_{\max}=32$, and the model is trained over three pyramid stages $\mathcal{R}=\{32,16,8\}$. 
Each input token contains four channels, including a relative 3D pivot coordinate and a discrete sign value, while the decoder predicts three coordinate channels and three sign logits for the classes $\{-1,0,+1\}$. 
The encoder uses hierarchical 3D convolutional stages with channel widths $[32,64,128,256]$, and the decoder mirrors this structure with widths $[512,256,128,64]$. 
Each stage contains two residual blocks, the bottleneck also contains two residual blocks, and layer normalization is used throughout the network. 
The latent code has $C_z=16$ channels, with the encoder predicting both posterior mean and log-variance.

\paragraph{Multi-stage training and optimization.}
Instead of optimizing all pyramid stages simultaneously, we randomly sample one stage at each iteration according to the weighted distribution $w_{32}:w_{16}:w_{8}=5:3:2$, corresponding to probabilities $0.5$, $0.3$, and $0.2$. 
This strategy emphasizes the finest stage while still exposing the model to coarser pyramid levels. 
When distributed training is used, the selected stage is synchronized across processes. 
The model is trained end-to-end with AdamW, using an initial learning rate of $5\times10^{-6}$ and cosine annealing over $500$ epochs. 
Training is performed in FP32 precision with gradient clipping at norm $1.0$, a block batch size of $64$, a mini-batch size of $2$, and a fixed random seed of $1234$.

\paragraph{Objective and validation.}
The Pyramid VAE is trained with a reconstruction-aware objective that supervises both pivot coordinates and orientation signs. 
Given a target token $x_i=(\mathbf{p}_i,s_i)$ and decoder prediction $\hat{x}_i=(\hat{\mathbf{p}}_i,\hat{\boldsymbol{\ell}}_i)$, the final objective is
\[
\mathcal{L}
=
\lambda_{\mathrm{xyz}}\mathcal{L}_{\mathrm{xyz}}
+
\lambda_{\mathrm{empty}}\mathcal{L}_{\mathrm{empty}}
+
\lambda_{\mathrm{sign}}\mathcal{L}_{\mathrm{sign}}
+
\lambda_{\mathrm{KL}}\mathcal{L}_{\mathrm{KL}},
\]
where
\[
\lambda_{\mathrm{xyz}}=1.0,\qquad
\lambda_{\mathrm{empty}}=0.01,\qquad
\lambda_{\mathrm{sign}}=0.5,\qquad
\lambda_{\mathrm{KL}}=10^{-4}.
\]
Here, $\mathcal{L}_{\mathrm{xyz}}$ is an $\ell_1$ coordinate reconstruction loss applied only to occupied voxels, encouraging accurate local pivot prediction. 
$\mathcal{L}_{\mathrm{empty}}$ penalizes non-zero coordinate predictions in empty voxels, preventing spurious surface evidence in inactive regions. 
$\mathcal{L}_{\mathrm{sign}}$ is a weighted three-class cross-entropy loss for the sign labels $\{-1,0,+1\}$, with class weights $(1.0,0.25,1.0)$ to balance occupied and empty states. 
$\mathcal{L}_{\mathrm{KL}}$ regularizes the latent posterior toward a standard Gaussian prior, enabling compact VAE-style latent modeling. 
Validation is performed every $5$ epochs using sampled validation batches, with both overall and per-stage metrics recorded; full reconstruction validation is optional and disabled in the reported configuration for memory efficiency.

\begin{table}[t]
\centering
\small
\caption{Architectural details of the Pyramid VAE encoder. 
The encoder starts from the input branch corresponding to the selected pyramid resolution $R\in\{32,16,8\}$ and progressively maps the token grid to a $4^3$ latent grid. 
Each ResBlock3D uses LayerNorm, SiLU activation, and $3\times3\times3$ convolutions.}
\label{tab:pvae_encoder}
\begin{tabular}{c|c|c|c}
\toprule
Stage & Input Resolution & Output Resolution & Block \\
\midrule
Input-$32$ 
& $32^3$ 
& $32^3$ 
& $\mathrm{Conv3D}(4,32,3)$ \\

$E_1$ 
& $32^3$ 
& $16^3$ 
& $\left[\mathrm{ResBlock3D}(32,32)\right]\times2,\ \mathrm{Down}(32,64)$ \\

Input-$16$ 
& $16^3$ 
& $16^3$ 
& $\mathrm{Conv3D}(4,64,3)$ \\

$E_2$ 
& $16^3$ 
& $8^3$ 
& $\left[\mathrm{ResBlock3D}(64,64)\right]\times2,\ \mathrm{Down}(64,128)$ \\

Input-$8$ 
& $8^3$ 
& $8^3$ 
& $\mathrm{Conv3D}(4,128,3)$ \\

$E_3$ 
& $8^3$ 
& $4^3$ 
& $\left[\mathrm{ResBlock3D}(128,128)\right]\times2,\ \mathrm{Down}(128,256)$ \\

Input-$4$ 
& $4^3$ 
& $4^3$ 
& $\mathrm{Conv3D}(4,256,3)$ \\

$E_4$ 
& $4^3$ 
& $4^3$ 
& $\left[\mathrm{ResBlock3D}(256,256)\right]\times2$ \\

Middle 
& $4^3$ 
& $4^3$ 
& $\left[\mathrm{ResBlock3D}(256,256)\right]\times2$ \\

Posterior Head 
& $4^3$ 
& $4^3$ 
& $\mathrm{LayerNorm},\ \mathrm{SiLU},\ \mathrm{Conv3D}(256,2C_z,3)$ \\
\bottomrule
\end{tabular}
\end{table}

\begin{table}[t]
\centering
\small
\caption{Architectural details of the Pyramid VAE decoder. 
The decoder maps the latent grid with $C_z=16$ channels to multi-resolution token predictions. 
Each output head predicts six channels, corresponding to three relative coordinate channels and three sign logits.}
\label{tab:pvae_decoder}
\begin{tabular}{c|c|c|c}
\toprule
Stage & Input Resolution & Output Resolution & Block \\
\midrule
Latent Input 
& $4^3$ 
& $4^3$ 
& $\mathrm{Conv3D}(C_z,512,3)$ \\

Middle 
& $4^3$ 
& $4^3$ 
& $\left[\mathrm{ResBlock3D}(512,512)\right]\times2$ \\

$D_4$ 
& $4^3$ 
& $4^3$ 
& $\left[\mathrm{ResBlock3D}(512,512)\right]\times2$ \\

Head-$4$ 
& $4^3$ 
& $4^3$ 
& $\mathrm{LayerNorm},\ \mathrm{SiLU},\ \mathrm{Conv3D}(512,6,3)$ \\

$D_3$ 
& $4^3$ 
& $8^3$ 
& $\mathrm{Up}(512,256),\ \left[\mathrm{ResBlock3D}(256,256)\right]\times2$ \\

Head-$8$ 
& $8^3$ 
& $8^3$ 
& $\mathrm{LayerNorm},\ \mathrm{SiLU},\ \mathrm{Conv3D}(256,6,3)$ \\

$D_2$ 
& $8^3$ 
& $16^3$ 
& $\mathrm{Up}(256,128),\ \left[\mathrm{ResBlock3D}(128,128)\right]\times2$ \\

Head-$16$ 
& $16^3$ 
& $16^3$ 
& $\mathrm{LayerNorm},\ \mathrm{SiLU},\ \mathrm{Conv3D}(128,6,3)$ \\

$D_1$ 
& $16^3$ 
& $32^3$ 
& $\mathrm{Up}(128,64),\ \left[\mathrm{ResBlock3D}(64,64)\right]\times2$ \\

Head-$32$ 
& $32^3$ 
& $32^3$ 
& $\mathrm{LayerNorm},\ \mathrm{SiLU},\ \mathrm{Conv3D}(64,6,3)$ \\
\bottomrule
\end{tabular}
\end{table}

\begin{table*}[t]
\centering
\small
\caption{Training hyperparameters of the Pyramid VAE.}
\label{tab:pvae_training_hparams}
\begin{tabular}{l|c}
\toprule
Hyperparameter & Value \\
\midrule
Dataset & ABO \\
Representation & Pyramid tokens \\
Global resolution & $512^3$ \\
Maximum block resolution $R_{\max}$ & $32$ \\
Training stages & $\{32,16,8\}$ \\
Stage sampling weights & $\{5.0,3.0,2.0\}$ \\
Input channels & $4$ \\
Output channels & $6$ \\
Latent channels $C_z$ & $16$ \\
Encoder channels & $[32,64,128,256]$ \\
Decoder channels & $[512,256,128,64]$ \\
Residual blocks per stage & $2$ \\
Middle residual blocks & $2$ \\
Normalization & LayerNorm \\
Batch size & $2$ \\
Block batch size & $64$ \\
Optimizer & AdamW \\
Learning rate & $5\times10^{-6}$ \\
Epochs & $500$ \\
Gradient clipping & $1.0$ \\
Precision & FP32 \\
Mixed precision & Disabled \\
KL weight & $10^{-4}$ \\
Sign CE weight & $0.5$ \\
Sign CE class weights & $[1.0,0.25,1.0]$ \\
XYZ loss weight & $1.0$ \\
Empty-space loss weight & $0.01$ \\
SDF cosine loss weight & $0$ \\
Validation interval & $5$ epochs \\
Checkpoint interval & $20{,}000$ steps \\
Maximum kept checkpoints & $10$ \\
Random seed & $1234$ \\
\bottomrule
\end{tabular}
\end{table*}

\clearpage
\section{Qualitative Visualization}
\label{app:qualitative}

\begin{figure*}[t]
    \centering
    \includegraphics[width=0.98\linewidth]{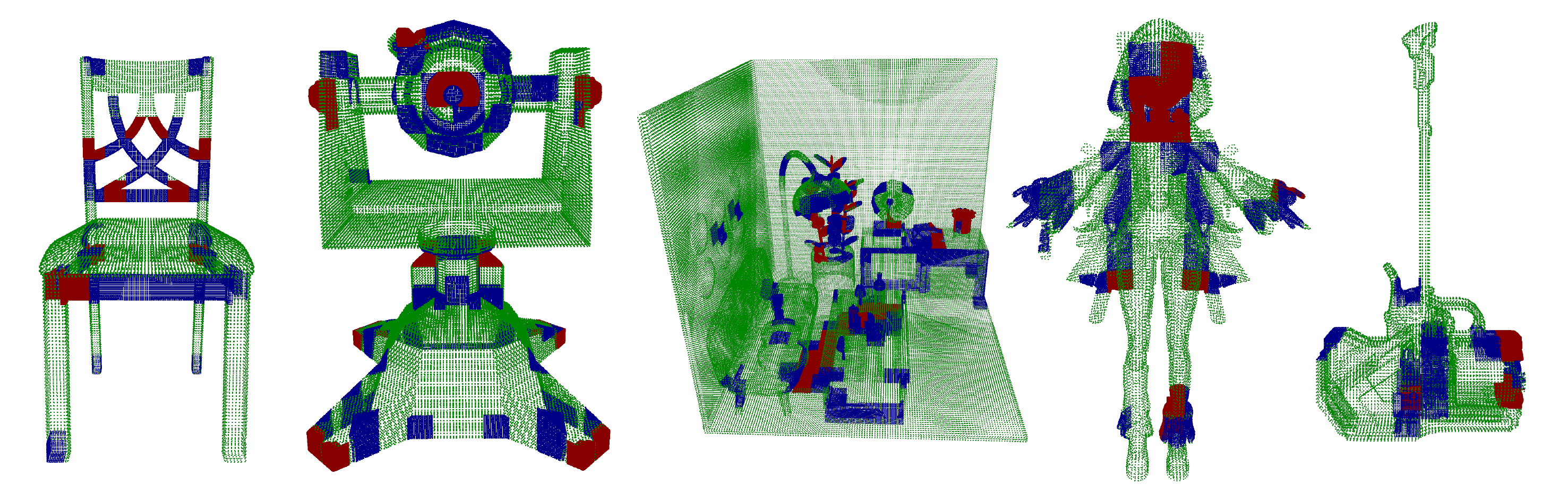}
    \caption{We visualize some pyramid token allocations across resolutions in Figure~\ref{fig:quali_pyramid}, where red/blue/green/cyan denote $R{=}512/256/128$, respectively. Lower-resolution tokens predominantly cover smooth and slowly varying regions, while higher-resolution tokens concentrate around silhouettes, sharp boundaries, and transition areas, demonstrating effective multi-scale budget allocation.}
    \label{fig:sampling_map}
\end{figure*}

\begin{figure*}[t]
    \centering
    \includegraphics[width=0.95\linewidth]{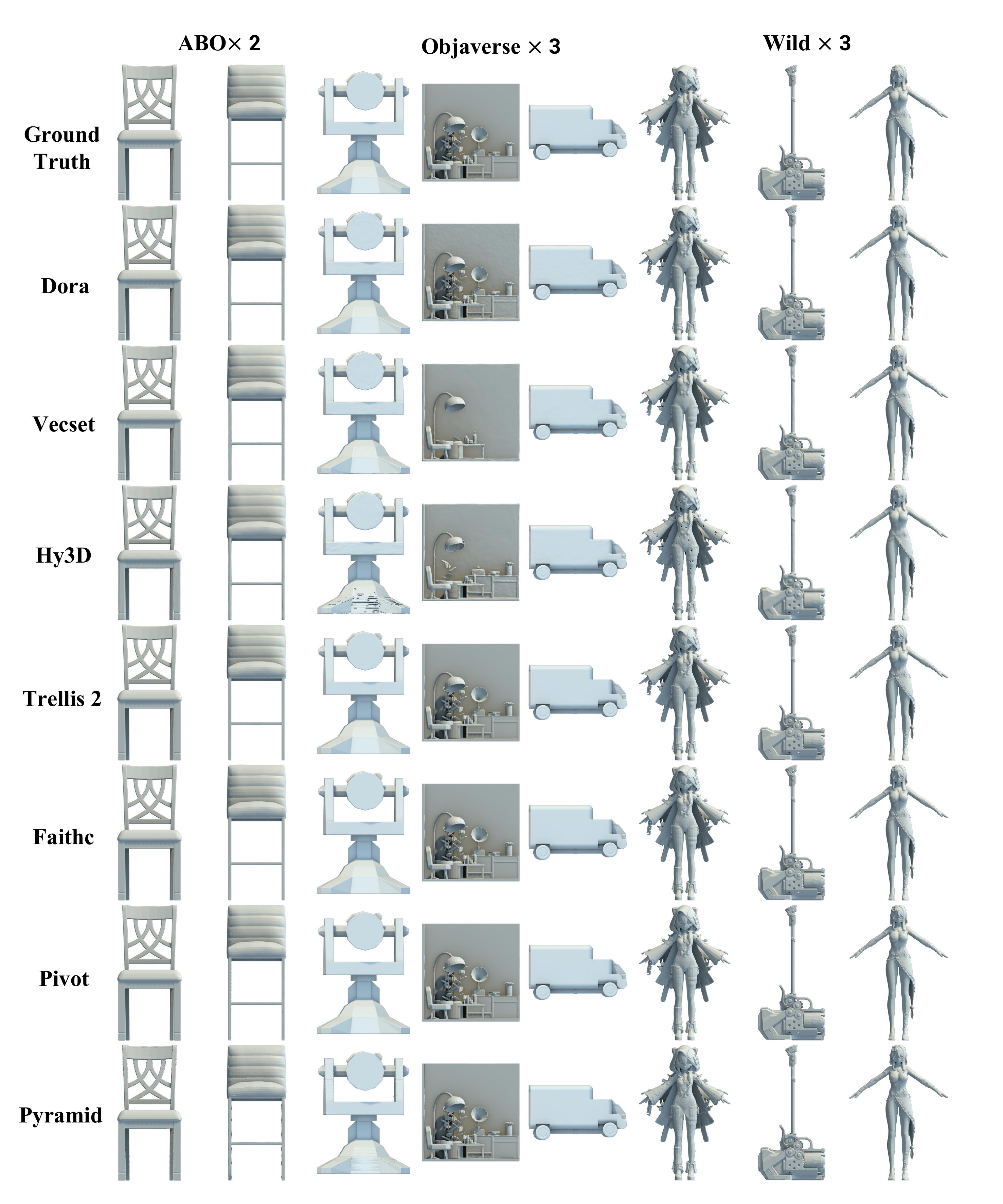}
    \caption{\textbf{Visualization results across different methods.} All comparisons are conducted purely at the mesh reconstruction level, without VAE tokenization or diffusion-based generation. With sufficiently dense sampling, all methods produce plausible reconstructions. Our P2Voxel achieves comparable quality with fewer active voxels (smaller token budget), and its minimal Dim~4 representation further benefits downstream token-based learning.}
    \label{fig:quali_pyramid}
\end{figure*}

\begin{figure*}[t]
    \centering
    \includegraphics[width=0.95\linewidth]{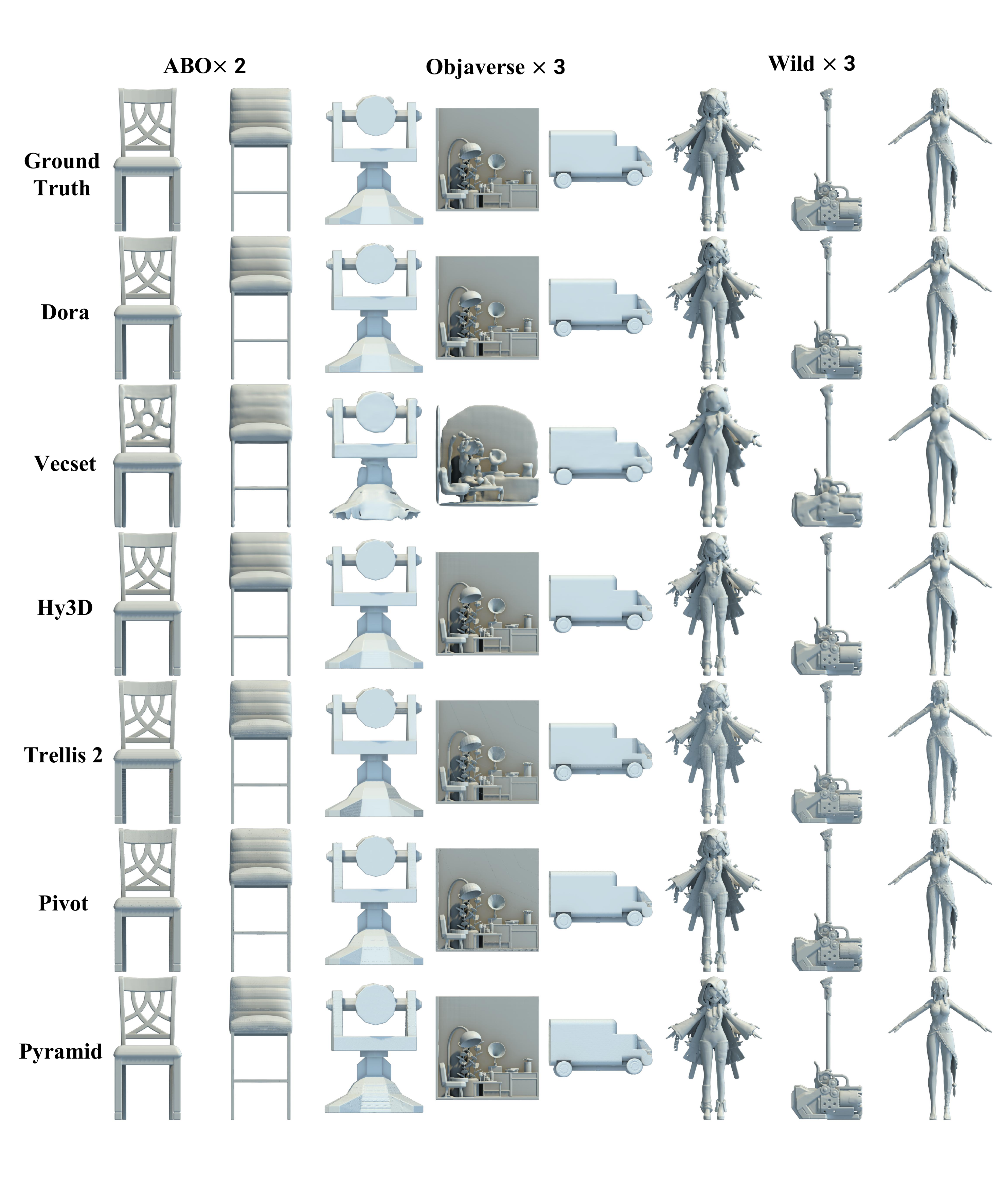}
    \caption{\textbf{Visualization results of VAE across different methods.} All comparisons are conducted purely at the mesh reconstruction level, without VAE tokenization or diffusion-based generation. With sufficiently dense sampling, all methods produce plausible reconstructions. Our P2Voxel achieves comparable quality with fewer active voxels (smaller token budget), and its minimal Dim~4 representation further benefits downstream token-based learning.}
    \label{fig:quali_vae}
\end{figure*}

We provide qualitative visualizations to further illustrate the behavior and reconstruction quality of P2Voxel.
Figure~\ref{fig:sampling_map} visualizes the pyramid sampling maps, showing how tokens are allocated across different resolutions according to local geometric complexity.
Figure~\ref{fig:quali_pyramid} compares mesh reconstruction results before VAE tokenization, focusing on the effectiveness of the pivot voxelization and pyramid sampling strategy itself.
Figure~\ref{fig:quali_vae} further presents reconstruction results after VAE compression, demonstrating that the compact pyramid latent representation preserves the main surface structure while supporting token-efficient learning.
Together, these visualizations show that P2Voxel allocates high-resolution tokens to geometrically informative regions and maintains high-quality reconstruction under both direct and VAE-compressed settings.

\section{Discussion}
\label{app:discuss}

\mypara{Future Work}
P2Voxel provides a compact and reconstruction-aware token space by converting dense mesh geometry into locally reconstructable pyramid pivot blocks. 
In this work, we mainly focus on geometry tokenization, deterministic reconstruction, and VAE-based compression, leaving full generative modeling as a natural next step. 
The pyramid sampling map can serve as a coarse spatial layout, while local pivot tokens describe fine-grained surface evidence within selected blocks, suggesting a coarse-to-fine pipeline that first predicts resolution allocation and then generates local pivot evidence. 
Another promising direction is to learn the sampling map directly from data. 
Although the current allocation strategy based on geometric scoring and fixed cut ratios already provides an effective efficiency--fidelity trade-off, a learned allocation module may better adapt the token budget to different object categories, surface complexity, and downstream tasks. 
Finally, more efficient CUDA-based kernels and parallel reconstruction procedures could make P2Voxel more suitable for large-scale 3D asset processing and real-time applications.

\mypara{Social Impact}
P2Voxel may have positive impact by reducing the storage, memory, and computational cost of high-resolution 3D geometry representation, benefiting applications such as 3D asset compression, digital content creation, simulation, robotics, virtual environments, and learning-based shape generation. 
By lowering the cost of representing detailed mesh geometry, the method may also make 3D learning pipelines more accessible to researchers and creators with limited computational resources. 
At the same time, more efficient 3D tokenization can make large-scale 3D asset synthesis and editing easier, especially when integrated into generative pipelines. 
Therefore, practical deployment should respect data provenance, copyright and licensing constraints of 3D assets, and responsible use of generated content. 
The present work focuses on geometry representation and reconstruction, and does not involve private data, human subjects, or high-risk deployment scenarios.

\end{document}